\documentclass[letterpaper]{article} 
\usepackage{aaai25}  
\usepackage{times}  
\usepackage{helvet}  
\usepackage{courier}  
\usepackage[hyphens]{url}  
\usepackage{graphicx} 
\usepackage{natbib}  
\usepackage{caption} 
\usepackage{floatrow}
\newfloatcommand{capbtabbox}{table}[][\FBwidth]
\usepackage{subfig}
\usepackage[lined,ruled]{algorithm2e}
\usepackage{amsmath}
\usepackage{amssymb}
\usepackage{mathtools}
\usepackage{amsthm}
\usepackage{newfloat}
\usepackage{listings}
\DeclareCaptionStyle{ruled}{labelfont=normalfont,labelsep=colon,strut=off} 
\floatstyle{ruled}
\newfloat{listing}{tb}{lst}{}
\floatname{listing}{Listing}
\title{Synergy-of-Thoughts: Eliciting Efficient Reasoning in Hybrid Language Models}
\author{
    Yu Shang\textsuperscript{1}\equalcontrib
    Yu Li\textsuperscript{2}\equalcontrib,
    Fengli Xu\textsuperscript{1},
    Yong Li\textsuperscript{1}
}
\affiliations{
    \textsuperscript{1}Tsinghua University, Beijing, China\\
    \textsuperscript{2}Harbin Institute of Technology, Shenzhen, China \\
    fenglixu@tsinghua.edu.cn,
    liyong07@tsinghua.edu.cn
}

\usepackage{bibentry}
\usepackage{booktabs}
\usepackage{multirow}
\begin{document}

\maketitle

\begin{abstract}
Large language models (LLMs) have shown impressive emergent abilities in a wide range of tasks, but the associated expensive API cost greatly limits the real application.
Previous works like chain-of-thought (CoT) and tree-of-thoughts (ToT) have predominately focused on enhancing accuracy, but overlook the rapidly increasing API cost, which could be particularly problematic for open-ended real-world tasks with huge solution spaces. 
Motivated by the dual process theory of human cognition, we propose ``Synergy of Thoughts'' (SoT) to unleash the synergistic potential of hybrid LLMs with different scales for efficient reasoning. 
By default, SoT uses smaller-scale language models to generate multiple low-cost intuitive thoughts, which resembles the parallel intuitions produced by \emph{System 1}. 
We then design a confidence evaluator where the intuitive thoughts are cross-evaluated and introduce a controllable threshold mechanism to decide their mutual conflict. 
If these intuitive thoughts exhibit conflicts, SoT will invoke the reflective reasoning of scaled-up language models to emulate the intervention of \emph{System 2}, which will override the intuitive thoughts and rectify the reasoning results.
This framework is model-agnostic and training-free, which can be flexibly implemented with various off-the-shelf LLMs.
Experiments on six representative reasoning tasks show that SoT substantially reduces the API cost by 38.3\%$\sim$75.1\%, and simultaneously achieves state-of-the-art reasoning accuracy and solution diversity. 
Notably, the average token cost reduction on open-ended tasks reaches up to 69.1\%.
\end{abstract}

%
\section{Introduction}
Initially conceived for autoregressive text generation, large language models (LLMs), such as GPT~\cite{brown2020language,radford2018improving,radford2019language} and PaLM~\cite{chowdhery2023palm}, have been shown to exhibit emergent abilities for reasoning tasks as they scale up~\cite{wei2022emergent}. 
A recent landmark study reveals LLMs can unlock their reasoning capability by employing ``Chain of Thought'' (CoT) ~\cite{wei2022emergent} prompts to produce intermediate steps for reasoning. 
The later ``Tree of Thoughts'' (ToT) framework~\cite{yao2023tree} further allows LLMs to deliberate on multiple reasoning paths and make high-quality global decisions via tree search. 
Search methods like ToT are believed to resemble the reflective reasoning mode found in human cognition, offering greater accuracy but at the expense of significantly high token costs paid for API services. 
For example, finding a solution for ``Game of 24'' with ToT consumes approximately 100 times more tokens compared to CoT~\cite{yao2023tree}. 
Besides, many open-ended real-world problems~\cite{zheng2023judging} with considerably larger solution spaces can also lead to high API costs.
Consequently, a critical research problem arises for practical LLM reasoning: Can we strike a more effective balance between reasoning accuracy and costs? 
This is significant for addressing reasoning problems in low-resource scenarios and facilitating the democratization of LLM reasoning. 
Some previous works~\cite{chen2023frugalgpt, yue2023large} propose to strategically choose a weaker or stronger LLM for solving the reasoning problem. Despite the cost reduction, these methods are unpromising to obtain higher performance than stronger LLMs, thus only achieving a discounted accuracy-cost trade-off.
For better accuracy-cost balance, it's challenging but promising to break down the reasoning process and design a more fine-grained and compact synergy mechanism for hybrid LLMs.

\begin{figure*}[t!]
\begin{center}
\centerline{\includegraphics[width=0.9\textwidth]{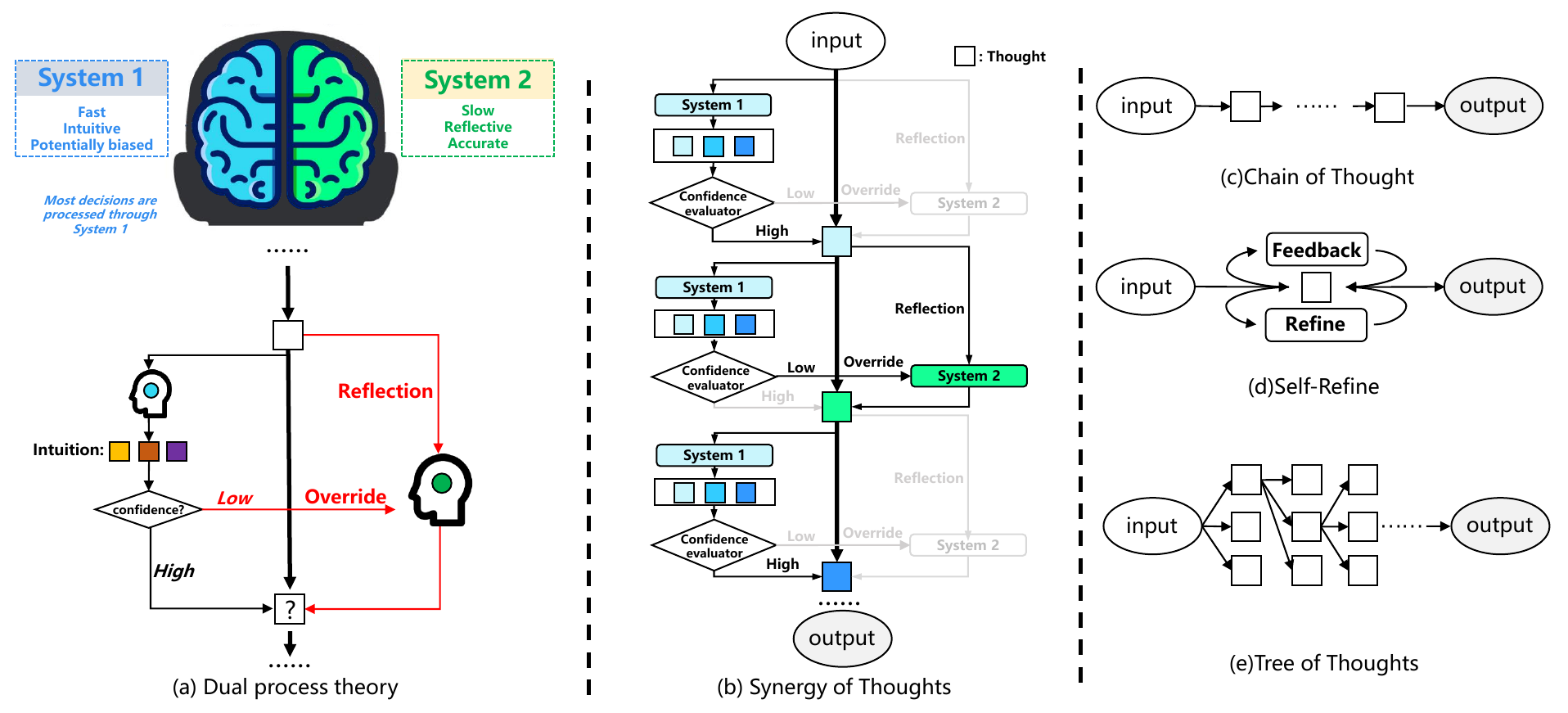}}
\caption{An illustration of dual process theory (a) and the main differences between SoT (b) and prior works (c) (d) (e). SoT is designed following the synergy paradigm of dual processes in human reasoning.}
\label{fig1}
\end{center}
\end{figure*}

Our study is motivated by human's cognition ability to efficiently tackle complex problems. The prevalent ``dual process'' theory suggests~\cite{evans2010intuition} there are two distinct systems in human reasoning: \emph{System 1}, capable of rapid, preconscious and intuitive responses; and \emph{System 2}, adept at high-effort, reflective reasoning. Research indicates that everyday decision-making is predominantly governed by \emph{System 1}~\cite{evans2010intuition}, providing fast responses with minimum resources through the intuition of associative experiences. Although \emph{System 1} makes accurate decisions most of the time, it is also identified as the source of various cognitive biases~\cite{kahneman2011thinking}, rendering it prone to errors if not properly monitored. On the contrary, \emph{System 2} can avoid intuitive biases through effortful reflection, which is widely encouraged in critical decision-making~\cite {croskerry2009clinical}.

Human reasoning has been observed to adopt a \emph{default-interventionist} mechanism, which reconciles these two competing systems by firstly using \emph{System 1} to generate low-effort \emph{default} responses, which may be \emph{intervened} upon with the reflection of high-effort \emph{System 2} if the confidence is low.  
Such mechanisms contribute to simultaneously enhancing both the reasoning accuracy and efficiency of humans.
Inspired by this, we propose ``Synergy of Thoughts'' (SoT) for efficient problem-solving with the synergy of hybrid LLMs (see Figure~\ref{fig1}). 
There are mainly three components in the framework of SoT: \emph{System 1}, \emph{System 2} and a confidence evaluator monitoring the synergy of the two systems. 
Firstly, to mimic the fast, low-effort \emph{System 1}, 
SoT employs several hybrid smaller-scale language models to emulate the intuitions derived from different associative experiences of humans.
Secondly, SoT implements the reflective \emph{System 2} with a scaled-up LLM which is considered to possess superior reasoning abilities.
Thirdly, we design a confidence evaluator to monitor the synergy of the two systems.
Specifically, it conducts a cross-evaluation of the intuitive thoughts from \emph{System 1}, generating a confidence score for each thought.
We then introduce a progressively increasing threshold and compare it with the highest confidence score to determine whether there are conflicts between intuitive thoughts.
Such a threshold control can also help flexibly adjust the workload of dual systems, delicately modulating the accuracy-cost balance in SoT. 
Regarding the whole workflow, for each reasoning step, SoT uses \emph{System 1} to generate multiple intuitive thoughts at low costs by default. 
Next, the confidence evaluator receives these intuitive thoughts and produces an intervention signal based on their conflicts.
If the intuitive thoughts are accepted, the final reasoning thought is selected as the best intuitive thought, otherwise, the reflective \emph{System 2} will be invoked to rectify and override the intuitive thoughts, ensuring faithfulness of the reasoning results.
With the above designs, SoT is expected to harness the synergistic potential of different-scale LLMs and deliver both efficient and accurate reasoning. 

Empirically, we conduct extensive experiments on six complex reasoning tasks, including both close-ended (Game of 24 \cite{yao2023tree}, Logic Grid Puzzle \cite{srivastava2022beyond}, GSM8K~\cite{cobbe2021training}) and open-ended problems (Trivia Creative Writing \cite{chen2023autoagents}, Open-ended QA \cite{chen2023autoagents}, Constrained Generation \cite{madaan2023self}). 
The results show that SoT achieves state-of-the-art reasoning accuracy on all six tasks. More importantly, it substantially reduces the API cost by 38.3\%$\sim$75.1\% compared to the second accurate baselines. 
Particularly, the token cost reduction in open-ended tasks (69.1\% on average) is higher than the close-ended tasks (42.6\% on average).   
Besides, we find SoT can also improve the solution diversity, probably because the \emph{default} module implemented by hybrid smaller-scale LLMs can access more diverse information sources. Our further analysis investigates the impact of intervention rate on the accuracy-cost balance, showing wide feasible implementations that could lead to a beneficial synergistic effect by using SoT.

In summary, our main contributions are as follows:
\begin{itemize}
    \item We propose SoT, a novel ``dual process'' theory-inspired framework that unleashes the synergistic potential of hybrid LLMs for cost-efficient reasoning. The framework is model-agnostic and can be implemented with various LLMs flexibly.
    \item We present a novel confidence evaluation mechanism for hybrid LLMs via cross-evaluation and threshold control, which can effectively monitor the faithfulness of the reasoning process.
    \item We conduct extensive experiments on six representative reasoning tasks. Empirical results demonstrate that SoT achieves state-of-the-art reasoning accuracy and solution diversity on all tasks. 
    More importantly, SoT substantially reduces the API cost by 38.3\% $\sim$ 75.1\% compared to the second accurate baseline. 
\end{itemize}

\begin{figure*}[!t]
\begin{center}
\centerline{\includegraphics[width=\textwidth]{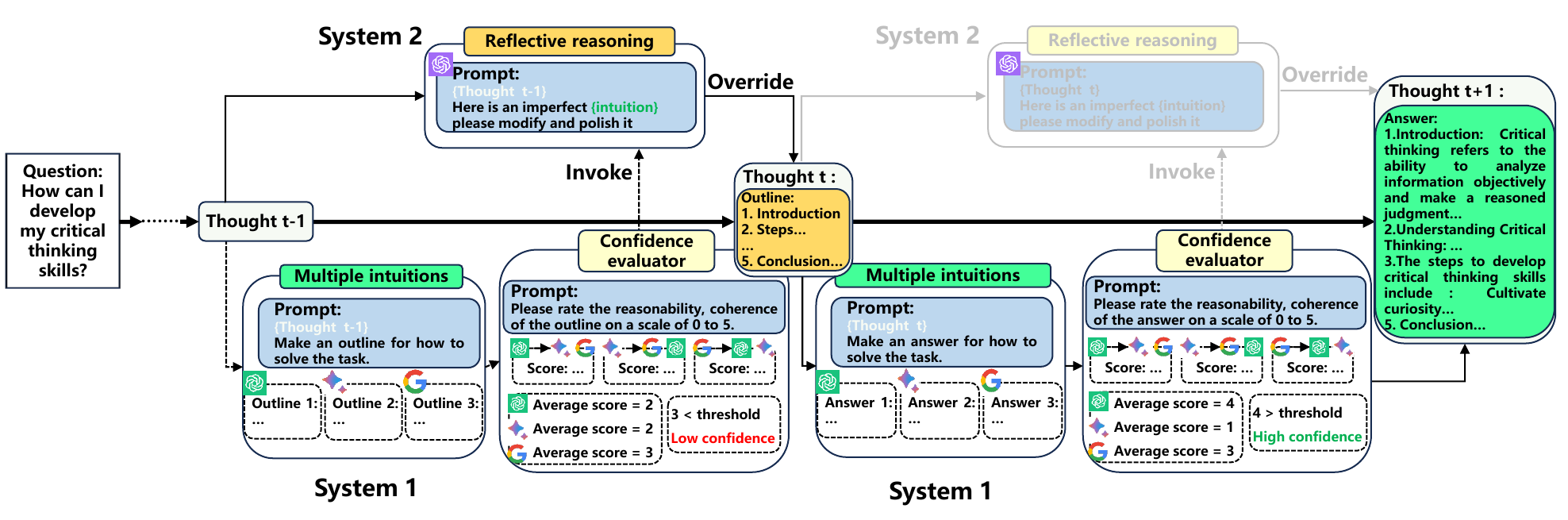}}
\caption{Overview of SoT illustrated with a two-step reasoning problem from the Open-ended QA task (making an outline in the first step and giving the answer in the second step). SoT prioritizes reasoning with default intuitions (System 1). When multiple intuitions are evaluated to be conflictual and low-confidence, SoT will intervene with reflective reasoning (invoking System 2) to override them.}
\label{overview}
\end{center}
\end{figure*}

\section{SoT: An Efficient Reasoning Framework with the Synergy of Hybrid LLMs} \label{method}
\textit{The evidence suggests that intuition is the dominant basis for real-world decision-making and is often effective; however, it also shows that reliance on intuition can be dangerous and that intervention with high-effort and explicit reasoning is often required, especially when problems have novel features.} --- Evans et al. \cite{evans2010intuition}

Motivated by the above default-interventionist theory, we introduce SoT, a framework that adaptively integrates two systems for both cost-efficient and accurate LLM reasoning. 
The whole framework of SoT is illustrated in Figure \ref{overview}, taking an open-ended QA problem as an example. 
Illustrations of more tasks and the complete SoT algorithm are shown in the appendix. 
Our framework provides a high-level design paradigm, which can have various model implementations in practice.
Next, we detail three key components in SoT, including System 1 and System 2, and the designed confidence evaluator for effective synergy of dual systems.
\subsection{Efficient Intuitive Thought Generation with System 1} 
System 1 is fast, intuitive, and largely dependent on relevant experiences, which is suitable to be implemented with smaller-scale language models that have not exhibited emergent reasoning abilities.
System 1 aims to efficiently draft intuitive thoughts and advance the reasoning process forward.
To mimic the diverse intuitions for humans, here we introduce $K$ distinct smaller-scale LLMs to generate diverse intuitive thoughts, denoted as $\{f_{Ii}(\cdot)|i \in \{1,2,...,K\}\}$.

Given the reasoning problem $p$, each LLM first proposes an initial thought independently, then we reconcile the competition of diverse intuitive thoughts via interactions to further refine the diverse intuitions. 
The detailed algorithm of System 1 is shown in the appendix and reasoning with System 1 at step $t$ can be formulated as:
\begin{equation}
    H_t = System \; 1(p_t;a_{t-1}) 
\end{equation}
where $H_t$ is the set of proposed intuitive thoughts, $p_t$ is the prompt of task description at the reasoning step $t$, $a_{t-1}$ denotes the thoughts of the last reasoning step.

\subsection{Reflective Thought Intervention with System 2}
System 2 is slow, reflective, and high-effort, which is expected to provide high-quality reasoning.
In the framework of SoT, System 2 is introduced to correct the biased intuitive thoughts from System 1 to ensure the quality of reasoning results.
To achieve this goal, we suggest implementing System 2 with scaled-up LLMs, because larger-scale LLMs exhibit more powerful reasoning capabilities than smaller-scale LLMs, which are promising to rectify the unfaithful intuitive thoughts.
Specifically, when intuitive thoughts are low-confidence, System 2 will be invoked for thought intervention. 
It takes the proposed intuitive thoughts at the current reasoning step as input and produces a rectified result overriding previous intuitive thoughts. 
Formally, given the prompt for reflective intervention $p_{ref}$ and the best intuitive thought $a_t$ at the reasoning step $t$, the reasoning process with System 2 is formulated as follows:
\begin{equation}
    a_{t} = System \; 2(p_{ref};a_t). 
\end{equation}

\subsection{Confidence Evaluation-based Thought Synergy}
Motivated by the synergy paradigm of dual processes of human reasoning \cite{evans2006heuristic,kahneman2002representativeness,stanovich2011rationality}, we follow a \textit{default-interventionist mechanism} to design the synergy framework of two competing systems.
It hypothesizes the two competing systems are reconciled by utilizing System 1 to obtain low-effort intuitive responses by default, which may be intervened upon with the reflection of high-effort System 2 when the confidence of intuitions is low.
Following this idea, in each reasoning step, SoT prioritizes utilizing System 1 to propose multiple intuitive thoughts $H$. However, these thoughts might be biased or hallucinated when facing novel and complex problems. When these intuitive thoughts show apparent conflicts, System 2 will be automatically invoked for intervention to rectify the reasoning process.
To provide high-quality signals, we propose a novel confidence evaluator for hybrid LLMs via cross-evaluation and threshold control.
\subsubsection{Confidence scoring with cross-evaluation}
We firstly leverage the diverse knowledge of hybrid LLMs to comprehensively measure the confidence of intuitive thoughts.
In detail, once System 1 provides $K$ intuitive thoughts $H = \{a_i | i \in \{1,2,...,K\}\}$, the $K$ LLMs in System 1 will conduct a cross-evaluation, where each intuitive thought is scored in turn by each LLM. 
Denote $p_{eval}$ as the prompt for confidence evaluation, the score of the i-th intuitive thought $a_i$ is formulated as: 
\begin{equation}
   V(a_i) = \frac{\sum_{j \in \{1,2,...,K\}} f_{Ij} (p_{eval}; a_i) }{K} 
\end{equation}
The larger score $V(a_i)$ indicates a more coherent evaluation toward the intuition $a_i$ and higher confidence. 
\subsubsection{Intervention signal generation with threshold control}
To obtain an executable evaluation criterion, we then introduce an adjustable threshold value $\varepsilon$.
The confidence evaluator will accept the highest confidential intuitive thought $a_k = argmax_{a \in H} V(a)$ if $V(a_k) > \varepsilon$, otherwise it will reject intuitive thoughts and invoke System 2 to overwrite the thoughts.
The intervention signal $p$ is generated according to the following rule:
    \begin{eqnarray}
		p =\left\{
		\begin{aligned}
			& True \quad V(a_k) > \varepsilon,  \; \\
			 & False \quad \text{otherwise}.
		\end{aligned}
		\right.
\end{eqnarray}
When the signal is True, System 2 will intervene and override the intuitive thoughts. The working frequency of System 1 and 2 can be easily controlled with varying threshold values. To further enhance the faithfulness of the reasoning pipeline, we progressively uplift the confidence threshold with the accumulated number of System 1-based reasoning steps. 
This is because the reasoning process with more intuitive thoughts is more likely to be biased, where the accepted threshold of intuitive thoughts should be raised.



Conceptually, SoT has several benefits via the harmonious synergy of two systems: (1) \textit{Cost efficiency.} Compared with existing advanced reasoning methods purely relying on high-cost System 2, SoT can significantly save token cost by using cost-efficient System 1 to propose intuitive thoughts. (2) \textit{Solution diversity.} SoT introduces diverse intuitions in System 1 to boost solution diversity, which is especially important for open-ended reasoning problems with huge solution space. (3) \textit{Competitive performance.} Although SoT introduces intuitive thoughts for reasoning, the default-interventionist mechanism can timely prevent the spread of bias and ensure the quality of reasoning results.  


\subsection{Theoretical Computation Cost Analysis} \label{theory}
To highlight the cost efficiency of SoT, we conduct a theoretical token cost analysis. 
For a more concise analysis, we assume that the output token cost of LLMs is proportional to the input token cost, thus do not distinguish input and output token prices. 
Denote the average API cost of every demonstration example in System 1 and System 2 as $C_I$ and $C_R$. 

In SoT, the API cost for System 1 with $K$ LLMs consists of three parts.
Firstly, the cost of initial thought generation of $K$ LLMs is $MKC_I$. 
Secondly, the API cost of interaction within $K$ LLMs consumes $(K-1)KC_I$.
Thirdly, each LLM will update their own thought based on feedback from other $K-1$ LLMs and the cost is $(K-1)KC_I$.
Therefore, the total cost of System 1 in SoT is:
\begin{equation}
    C_{system1} = (M+2K-2)KC_I, 
\end{equation}
where $M$ denotes the number of demonstrations in the prompt. 
The cost of the confidence evaluation in SoT comes from the cross-evaluation of $K$ LLMs in System 1, and the total of this part is: 
\begin{equation}
    C_{eval} = (K-1)KC_I,
\end{equation}
 If the confidence of intuitions is low, System 2 will be invoked with only the highest confidential intuitive thought as input, thus the cost is:
 \begin{equation}
    C_{system2} = C_R.
\end{equation}
By comparison, the cost of reasoning with only reflective System 2 is:
 \begin{equation}
    C' = MC_R.
\end{equation}
Denote the intervention rate as $r$, when $(1-r)(C_{system1}+C_{eval})+r(C_{system1}+C_{eval}+C_{system2}) < C'$ is satisfied, SoT is expected to effectively save API costs and the corresponding condition of $r$ is (taking $M=1$ in most cases):
 \begin{equation}
    r<1-\frac{(3K-2)KC_I}{C_R}.
    \label{eq9}
\end{equation}
In the experiment part, we present a detailed analysis combined with specific cost statistics of used LLMs.

\setlength{\tabcolsep}{1mm} 
\begin{table*}[t!]
    \centering
    \begin{tabular}{ccccccccccccccc}
    \hline
    \multirow{2}{*}{Method} & \multicolumn{4}{c}{Game of 24} & \multicolumn{4}{c}{Trivia Creative Writing} & \multicolumn{3}{c}{Logic Grid Puzzle} & \multicolumn{3}{c}{GSM8K} \\
    \cmidrule(lr){2-5} \cmidrule(lr){6-9} \cmidrule(lr){10-12} \cmidrule(lr){13-15}
         & Acc & Div & Cost & TFLOPS & Acc & Div & Cost & TFLOPS & Acc & Cost & TFLOPS & Acc & Cost & TFLOPS \\ \hline
        CoT(best of 1) & 4\% & 1.0 & 0.87 & 5.22E+4 & 67.1\% & 3.8 & 3.37 & 2.02E+5 & 65.8\% & 6.72 & 4.03E+5 & 87.8\% & 9.58 & 5.75E+5 \\ 
        CoT(best of 5) & 14\% & 1.1 & 1.73 & 1.04E+5 & 73.4\% & 3.9 & 13.09 & 7.86E+5 & 67.1\% & 27.26 & 1.64E+6 & 91.3\% & 35.78 & 2.15E+6 \\ 
        LLM-cascade & 8\% & 1.0 & 1.24 & 7.86E+4 & 65.7\% & 5.1 & 3.21 & 1.90E+5 & 62.1\% & 7.98 & 5.11E+5 & 89.1\% & 7.75 & 4.39E+5 \\ 
        Self-refine & 20\% & 1.2 & 24.83 & 1.51E+6 & 78.2\% & 4.9 & 17.79 & 1.07E+6 & 60.6\% & 33.37 & 2.00E+6 & 91.1\% & 23.45 & 1.41E+6 \\ 
        ToT & 64\% & 2.1 & 23.37 & 1.40E+6 & 76.8\% & 4.4 & 27.32 & 1.64E+6 & 66.1\% & 38.66 & 2.32E+6 & 91.8\% & 25.53 & 1.53E+6 \\ 
        SPP & 12\% & 1.2 & 29.97 & 1.80E+6 & 79.9\% & 5.8 & 10.94 & 6.56E+5 & 68.3\% & 20.68 & 1.24E+6 & 84.6\% & 63.72 & 3.82E+6 \\ 
        MAD+judge & 22\% & 1.3 & 28.09 & 1.69E+6 & 77.4\% & 6.1 & 17.00 & 1.02E+6 & 66.8\% & 45.00 & 2.70E+6 & 89.3\% & 43.68 & 2.62E+6 \\ \hline
        $SoT_O$ & \underline{73\%} & \underline{2.3} & 13.14 & 1.07E+6 & \underline{82.2\%} & \textbf{6.5} & 2.96 & 1.82E+5 & \underline{69.9\%} & 11.58 & 7.91E+5 & \underline{93.4\%} & 12.91 & 8.74E+5 \\ 
        $SoT_C$ & \textbf{76\%} & \textbf{2.4} & 14.42 & 1.28E+6 & \textbf{83.1\%} & \underline{6.3} & 3.41 & 2.38E+5 & \textbf{71.5\%} & 12.23 & 8.93E+5 & \textbf{94.0\%} & 13.14 & 8.98E+5 \\ \hline
    \end{tabular}
    {\caption{Results on Game of 24, Trivia Creative Writing, Logic Grid Puzzle and GSM8K tasks.}
    \label{tab:main1}}
\end{table*}

\begin{table}[t!]
    \begin{tabular}{lllll}
    \hline
        Methods & FairEval & Div & Cost & TFLOPS \\ \hline
        $SoT_O$ & ~ & \textbf{6.1} & 7.94 & 5.01E+05 \\ \hline
        v.s. CoT(best of 1) & 71.4\% & 4.2 & 4.84 & 2.91E+05 \\
        v.s. CoT(best of 5)  & 63.7\% & 4.2 & 20.11 & 1.21E+06 \\
        v.s. LLM-cascade & 73.9\% & 5.4 & 4.39 & 2.60E+05 \\
        v.s. Self-refine & 58.4\% & 5.3 & 31.86 & 1.91E+06 \\ 
        v.s. ToT & 65.2\% & 4.7 & 58.52 & 3.51E+06 \\ 
        v.s. SPP & 68.6\% & 4.6 & 26.69 & 1.60E+06 \\ 
        v.s. MAD+judge & 59.9\% & \underline{5.6} & 36.59 & 2.20E+06 \\ \hline
    \end{tabular}
    {\caption{Results on Constrained Generation task (FairEval value larger than 50\% means results from $SoT_O$ are better).}
    \label{tab:main5}}
\end{table}

\section{Experiments} \label{exp}
\subsection{Experimental Settings}
\textbf{Task setup.}
We evaluate SoT and compared methods on six representative reasoning tasks including three close-ended tasks (Game of 24 \cite{yao2023tree}, Logic Grid Puzzle \cite{srivastava2022beyond}, GSM8K~\cite{cobbe2021training}) and three open-ended tasks (Trivia Creative Writing \cite{chen2023autoagents}, Open-ended QA \cite{chen2023autoagents}, Constrained Generation \cite{madaan2023self}). For each task, we use the same number of demonstration examples with original papers (one-shot in Game of 24 and zero-shot in others). 

\noindent\textbf{Baselines.}
We compare SoT with the following competitive baselines of LLM reasoning (all methods are implemented with GPT-4 unless otherwise stated, for a more fair performance comparison):
\begin{itemize}
    \item \textbf{Chain-of-thought (CoT)} \cite{wei2022chain}: It firstly proposes to guide LLMs to think step-by-step for reasoning. For a fair comparison, we conduct multiple trials until reaching a similar token cost of our method. For example, the result of CoT (best of 5) is reported as the best performance among five independent trials of CoT.  
    \item \textbf{Self-refine} \cite{madaan2023self}: It iteratively produces self-feedback and refines the results.
    The maximum refinement round is set as 4.
    \item \textbf{Tree-of-thoughts (ToT)} \cite{yao2023tree}: It generates multiple thought paths and searches the best solution with the heuristics method.
    We set the candidate number at each step as 5.
    \item \textbf{LLM-cascade} \cite{yue2023large}: It designs a dynamic reasoning framework with weaker and stronger LLMs controlled by checking the answer consistency of the weak LLM. 
    Different from our method, it only utilizes weaker LLMs from a single source and disentangles the thoughts of weaker and stronger LLMs, limiting the performance upper bound.
    We follow the implementation of CoT-2D-Vote in the original paper, setting the number of sampling paths as 20 for GPT-3.5 and 3 for GPT-4.
    \item \textbf{SPP} \cite{wang2023unleashing}: It transforms a single LLM into different personas and lets them collaborate to solve reasoning problems. We use GPT-4 for the implementation.
    \item \textbf{Multi-agent debate with judgment (MAD+judge)} \cite{liang2023encouraging}: It designs a multi-agent debate pipeline with judgment for reasoning problems. 
    We set three agents implemented with GPT-4 and three rounds of debate.
\end{itemize}

\begin{table}[t!]
    \begin{tabular}{lllll}
    \hline
        Methods & FairEval & Div & Cost & TFLOPS \\ \hline
        $SoT_O$ & ~ & \textbf{5.5} & 6.23 & 4.48E+05 \\ \hline
        v.s. CoT(best of 1) & 67.2\% & 3.1 & 2.27 & 1.36E+05 \\ 
        v.s. CoT(best of 5) & 63.9\% & 3.3 & 8.72 & 5.23E+05 \\
        v.s. LLM-cascade & 71.1\% & 4.5 & 2.49 & 1.50E+05 \\
        v.s. Self-refine & 58.6\% & 4.2 & 15.27 & 9.16E+05 \\ 
        v.s. ToT & 60.8\% & 3.3 & 19.44 & 1.17E+06 \\ 
        v.s. SPP & 66.1\% & 3.8 & 8.33 & 5.00E+05 \\ 
        v.s. MAD+judge & 55.2\% & \underline{4.7} & 17.00 & 1.02E+06 \\  \hline
    \end{tabular}
    {
    \caption{Results on Open-ended Question Answer task (FairEval value larger than 50\% means results from $SoT_O$ are better).}
    \label{tab:main6}}
\end{table}

\noindent\textbf{Metrics.} We evaluate the methods for reasoning tasks from four perspectives: 
\begin{itemize}
    \item \textbf{Accuracy (Acc)}: For three close-ended tasks, accuracy is measured directly by the generated answer and ground truths. For Trivia Creative Writing task, accuracy is calculated by \# correct answer mentions/\# trivia questions. For other open-ended tasks, we utilize FairEval \cite{wang2023large} to test the answer quality, following prior works \cite{chen2023autoagents,chan2023chateval}. 
    \item \textbf{Diversity (Div)}: Solution diversity of content generated by LLMs has long been an important concern \cite{kirk2023understanding,padmakumar2023does}, which is also important for reasoning tasks, especially open-ended problems with huge solution spaces. For Game of 24, we use the number of generated correct answers to measure solution diversity. For Logic Grid Puzzle and GSM8K, there's no concept of diversity. For three open-ended tasks, we modify the prompt of FairEval to let it give a diversity score (from 1 to 10) of two answers from the same model. 
    \item \textbf{API cost (Cost)}: It records the dollar cost of running the method once using API services, with great attention in prior works \cite{yue2023large,yao2023tree}.
    \item \textbf{TFLOPS}: It reflects the computational complexity, which is estimated according to the number of parameters following \cite{kaplan2020scaling}. 
\end{itemize}

\noindent\textbf{SoT setup.}
SoT provides a model-agnostic framework, which is flexible and has various implementations. 
In our experiments, we try two representative implementations, named $SoT_O$ and $SoT_C$. 
In $SoT_O$, we implement System 1 with three popular open-source and small-scale LLMs including Mistral-7B~\cite{jiang2023mistral}, LLaMA-13B~\cite{touvron2023llama} and Yi-34B~\cite{young2024yi}. GPT-4~\cite{achiam2023gpt} is chosen to implement the intervention with System 2. 
In $SoT_C$, we implement System 1 with three closed-source and relatively smaller-scale LLMs including GPT-3.5~\cite{achiam2023gpt}, PaLM2~\cite{anil2023palm} and Gemini1pro~\cite{team2023gemini}. GPT-4 is chosen to implement the intervention with System 2 as before. 
Besides, it's practicable to implement SoT with other LLMs. 
We set the threshold value $\varepsilon$ as 3.5 and the progressive increasing rate as 10\% in the confidence evaluation for all tasks (more choices are analyzed in the experiment part). All LLMs are accessed via APIs and we run all experiments on a CPU machine with 16GB memory. 

\begin{table}[t!]
    \begin{tabular}{lllll}
    \hline
        Methods & Acc & Div & Cost & TFLOPS \\  \hline
        SoT (default)  & \textbf{73\%} & \textbf{2.4} & 13.14 & 1.07E+06 \\
        SoT (3 LLaMA-13B)  & 61\% & 1.8 & 14.45 & 1.19E+06 \\
        SoT (3 Mistral-7B) & 64\% & \underline{2.0} & 13.86 & 1.12E+06 \\ 
        SoT (3 Yi-34B)   & \underline{67\%} & \underline{2.0} & 14.20 & 1.18E+06 \\ 
        SoT (1 LLaMA-13B)  & 55\% & 1.5 & 14.92 & 1.26E+06 \\ 
        SoT (1 Mistral-7B)  & 60\% & 1.8 & 13.98 & 1.19E+06 \\   
        SoT (1 Yi-34B) & 58\% & 1.7 & 14.71 & 1.21E+06 \\ 
        \hline
    \end{tabular}
    {\caption{Performance of different model choices for $SoT_O$ on Game of 24.}
    \label{tab:ab-open-24}}
\end{table}

\begin{table}[t!]
        \begin{tabular}{lllll}
    \hline
        Methods & Acc & Div & Cost & TFLOPS \\  \hline
        SoT (default)  & \textbf{76\%} & \textbf{2.4} & 14.42 & 1.28E+06 \\  
        SoT (3 GPT-3.5)  & \underline{70\%} & \underline{2.2} & 15.65 & 1.35E+06 \\ 
        SoT (3 PaLM2) & 67\% & 2.0 & 16.93 & 1.50E+06 \\
        SoT (3 Gemini1pro)  & 69\% & 2.1 & 15.34 & 1.28E+06 \\
        SoT (1 GPT-3.5)  & 67\% & 2.1 & 13.73 & 1.12E+05 \\
        SoT (1 PaLM2)  & 59\% & 1.9 & 12.88 & 1.02E+06 \\   
        SoT (1 Gemini1pro) & 62\% & 1.9 & 12.12 & 9.63E+05 \\ 
        \hline
    \end{tabular}
    {
    \caption{Performance of different model choices for $SoT_C$ on Game of 24.}
    \label{tab:ab-close-24}}
\end{table}

\subsection{Main Results}
We present the performance comparison of SoT and baselines on six representative reasoning tasks in Table \ref{tab:main1}, \ref{tab:main5} and \ref{tab:main6}. From the results, we have the following observations:

\noindent\textbf{(1) SoT achieves the best reasoning accuracy with significantly reduced computation cost across different tasks.}
Broadly, SoT outperforms all compared methods in terms of reasoning accuracy (or quality evaluation with FairEval). 
For three close-ended reasoning tasks, compared with the best baseline, on average SoT improves 8.6\% reasoning accuracy, simultaneously saving 42.6\% token costs and 30.0\% TFLOPS.
For three open-ended reasoning tasks, compared with the best baseline, on average SoT improves 5.9\% reasoning accuracy, simultaneously saving 69.1\% token costs and 64.5\% TFLOPS.
We also provide intuitive comparisons of reasoning accuracy and solution diversity versus token costs/TFLOPS on Game of 24 and Trivia Creative Writing in the appendix. 
Overall, SoT achieves the best trade-off between reasoning performance and cost efficiency.

\noindent\textbf{(2) SoT benefits solution diversity.} Except for reasoning accuracy, we also pay attention to the solution diversity of reasoning tasks, which is especially vital for some open-ended problems. It can be found that solutions generated by SoT possess the highest diversity among all methods on four tested tasks. Specifically, for close-ended and open-ended tasks, SoT achieves 14.3\% and 12.7\% solution diversity improvement on average compared with the best baseline. This might be attributed to the integration of diverse intuitions in System 1 and the further synergy of dual systems. 

\noindent\textbf{(3) SoT achieves superior performance under various implementations.} 
We implement SoT with two versions including both open-source and closed-source LLMs for System 1. From the results, SoT consistently outperforms baselines in terms of the trade-off between reasoning performance and costs. This demonstrates the superiority of our designed framework itself and verifies the flexibility of implementations for SoT.

\begin{table}[t!]
     \begin{tabular}{lllll}
    \hline
        Methods & Acc & Div& Cost & TFLOPS \\  \hline
        SoT (default)  & \textbf{82.2\%} & \textbf{6.5} & 2.96 & 1.82E+05 \\ 
        SoT (3 LLaMA-13B)  & 77.1\% & \underline{5.8} & 3.31 & 2.01E+05 \\ 
        SoT (3 Mistral-7B) & 79.6\% & 5.6 & 3.22 & 1.97E+05 \\
        SoT (3 Yi-34B)  & \underline{80.3\%} & 5.7 & 3.56 & 2.13E+05 \\
        SoT (1 LLaMA-13B)  & 75.3\% & 4.4 & 3.69 & 2.16E+06 \\
        SoT (1 Mistral-7B)  & 76.2\% & 4.5 & 2.88 & 1.71E+06 \\   
        SoT (1 Yi-34B) & 77.8\% & 4.2 & 3.27 & 1.96E+06 \\ 
        \hline
    \end{tabular}
    {\caption{Performance of different model choices for $SoT_O$ on Trivia Creative Writing.}
    \label{tab:ab-open-writing}}
\end{table}

\begin{table}[t!]
        \begin{tabular}{lllll}
    \hline
        Methods & Acc & Div & Cost & TFLOPS \\  \hline
        SoT (default) & \textbf{83.1\%} & \textbf{6.3} & 3.41 & 2.38E+05 \\  
        SoT (3 GPT-3.5)  & \underline{80.9\%} & 5.7 & 3.94 & 2.60E+05 \\  
        SoT (3 PaLM2) & 78.6\% & \underline{6.0} & 3.77 & 2.51E+05 \\
        SoT (3 Gemini1pro)  & 80.1\% & 5.8 & 3.38 & 2.32E+05 \\
        SoT (1 GPT-3.5)  & 78.4\% & 4.2 & 3.11 & 2.11E+05 \\ 
        SoT (1 PaLM2)  & 75.9\% & 4.5 & 3.32 & 2.20E+05 \\    
        SoT (1 Gemini1pro) & 77.3\% & 4.5 & 3.02 & 2.03E+05 \\
        \hline
    \end{tabular}
    {
    \caption{Performance of different model choices for $SoT_C$ on Trivia Creative Writing.}
    \label{tab:ab-close-writing}}
\end{table}

\subsection{In-depth Analysis of SoT} \label{in-depth}
\textbf{Study of model choices.} 
In this part, we explore the impact of choosing different LLMs for the implementation of SoT including both open-source and closed-source LLMs. Here we show the results on Game of 24 and Trivia Creative Writing in Table \ref{tab:ab-open-24}, \ref{tab:ab-close-24}, \ref{tab:ab-open-writing}, \ref{tab:ab-close-writing}. 
For each version of SoT, we have tried implementations with each single LLM and hybrid LLMs.
From the results, it can be found that utilizing hybrid LLMs to propose intuitions can benefit both accuracy and solution diversity. 
This might be attributed to hybrid LLMs' ability to access more diverse information sources.
Besides, utilizing more LLMs in System 1 can also harvest performance gain. 

\begin{figure}[t!]
\begin{center}
\centerline{\includegraphics[width=\textwidth]{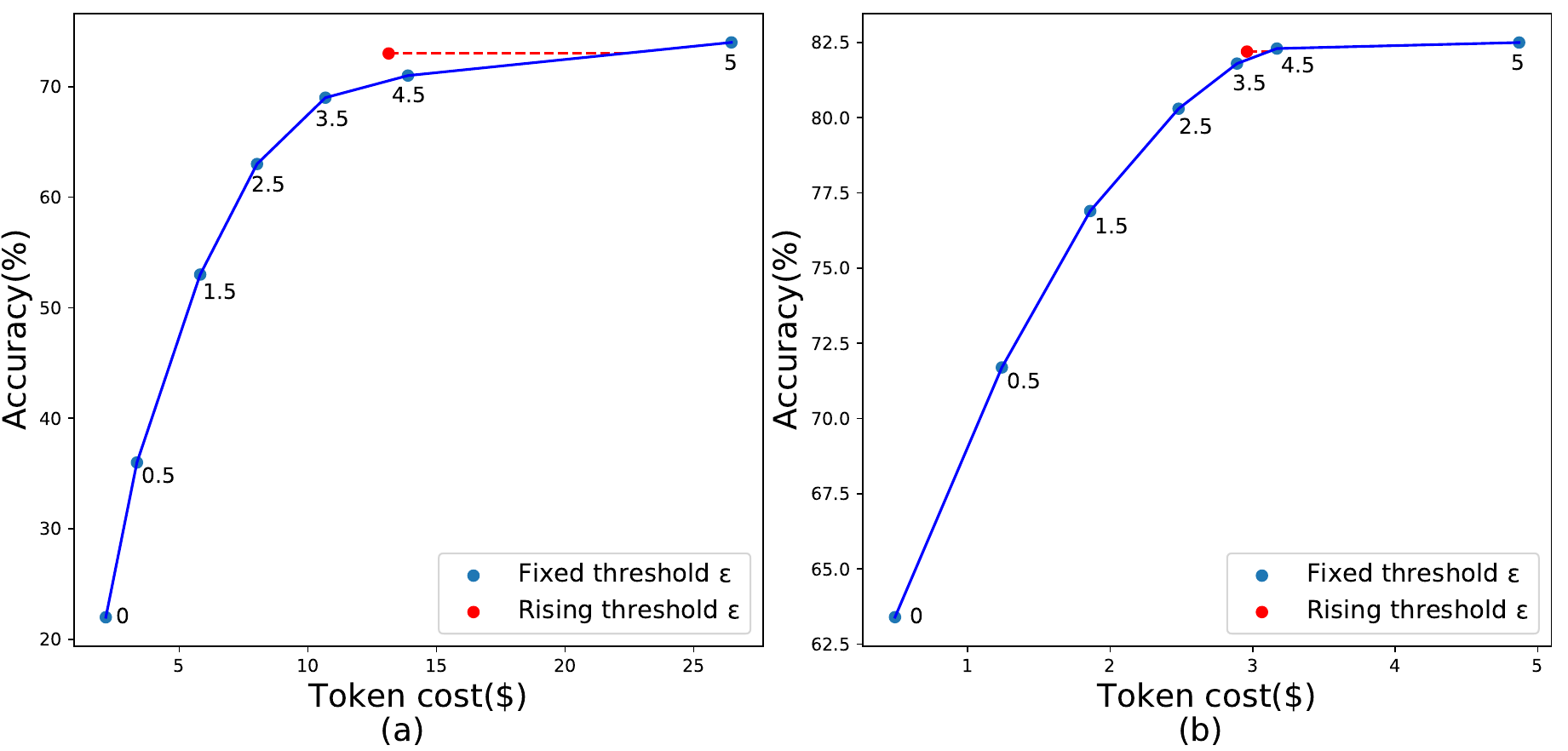}}
\caption{Reasoning cost-accuracy trade-off under different threshold value choices in $SoT_O$ on (a) Game of 24 and (b) Trivia creative writing. The number in the figure means the chosen threshold value.}
\label{fig_thr}
\end{center}
\end{figure}

\noindent\textbf{Study of the threshold value in confidence evaluation.} 
A key design of SoT is the confidential threshold $\varepsilon$ to adjust the workload of System 1 and System 2. In our experiments, we regulate that the confidence score ranges from 0 to 5. We then try 7 different threshold values within this interval for $SoT_O$ and present the cost-accuracy trade-off on three tasks in Figure \ref{fig_thr}. Specifically, $\varepsilon=0$ means intuitive thoughts from System 1 will always be accepted and SoT degrades to pure System 1, which is cost-efficient but inaccurate.  $\varepsilon=5$ means intuitive thoughts from System 1 will always be overwritten by System 2 and SoT becomes a pure System 2, which is accurate but costly. Observing that the incremental accuracy gain becomes much weaker after $\varepsilon>3.5$, for simplicity, we set $\varepsilon=3.5$ in all experiments. Besides, we introduce a progressive threshold-rising strategy where $\varepsilon$ increases 10\% each time from 3.5 with the accumulation of intuition-based reasoning steps. This further enhances reasoning performance (shown as red dots in Figure \ref{fig_thr}) and mitigates bias propagation in the reasoning process.

\noindent\textbf{Feasible intervention rate for efficient LLM synergy.} 
Here we conduct a further study combined with empirical statistics to show the practical options of efficient LLM synergy. 
In terms of $SoT_C$, the most diverse LLM combination (GPT-3.5/PaLM2/Gemini1pro + GPT-4) obtains the lowest intervention rate on the whole. 
We calculate the upper bound of the required intervention rate of $SoT_C$ according to Eq.(\ref{eq9}) and token costs shown in the appendix. The required condition is $r<57.2\%$, which is higher than the empirical value. 
By comparison, some more homogeneous synergy groups such as 3 PaLM2 + GPT-4 come with higher intervention rates. 
Similarly, for $SoT_O$ implemented with LLaMA-13B/Mistrial-7B/Yi-34B + GPT-4, it's expected to save token costs only if $r<86.3\%$.
Such estimation can be easily generalized to any LLM combination and provides a quick assessment of the feasibility of efficient LLM synergy.

\section{Related Work}
\textbf{Reasoning with LLMs.} With the blooming of LLMs, there has been plenty of work utilizing LLMs to address reasoning problems \cite{wei2022chain,yao2023tree,wang2022self,zhou2022least,zhang2023cumulative}. 
The early method is to add few-shot examples in the prompt and let LLMs answer the target question, \textit{i.e.,} standard IO prompting \cite{brown2020language}. However, the resultant performance is very limited, then some more advanced prompting methods are proposed to facilitate the reasoning ability of LLMs. 
For example, CoT encourages LLMs to think step-by-step, which can activate their inherent reasoning abilities \cite{wei2022chain}. 
ToT further explores multiple different thought paths and searching in the thought tree with heuristics methods \cite{yao2023tree}.
Besides, there are also other advanced mechanisms introduced to improve LLM reasoning abilities such as reflection \cite{shinn2023reflexion} and refinement \cite{madaan2023self}.
Another line of work to enhance the reasoning ability of LLMs is to develop a multi-LLM collaboration system \cite{du2023improving,liang2023encouraging,wang2023unleashing,chen2023reconcile,sun2023corex,yin2023exchange}. 
Although the above methods facilitate LLM reasoning performance, they all improve reasoning performance along with higher API costs. 
Differently, our method explores the effective synergy of the dual systems for a better balance between reasoning performance and cost efficiency.

\textbf{Cost-efficient reasoning with LLMs.}
Efficiency and cost are critical challenges for LLM reasoning due to the involved complex computations. To improve the speed and cost-effectiveness of LLM reasoning, there have been several approaches, such as quantization \cite{tao2022compression} and model pruning \cite{sun2023simple}.
Besides, some works focused on how to utilize the paid API efficiently \cite{chen2023frugalgpt,vsakota2023fly}. For example, Chen et al. \cite{chen2023frugalgpt} proposed a framework that sends the query to a series of LLMs sequentially if the answers given by the prior model are considered unacceptable.
However, all the above methods need to transform the model itself or introduce external fine-tuned verifiers, bringing additional computation costs. Different from the previous model-side modification, we explore a novel path for reducing reasoning cost via diverse LLM synergy, which is training-free and general purpose. 

\section{Conclusion}
We introduce SoT, an effective hybrid LLM synergy framework for efficient reasoning without any additional training or fine-tuning. Following the default-interventionist mechanism of human decision-making, SoT can adaptively switch between intuitive and reflective thoughts, thus facilitating a better balance between reasoning performance and computation costs. Extensive experiments on broad reasoning tasks emphasize the superiority and generalizability of our method. Compared with the best baseline, SoT can further enhance reasoning accuracy and solution diversity, simultaneously reducing the API cost by 38.3\% $\sim$ 75.1\%. We hope that this work can provide a novel perspective for efficient LLM reasoning with model synergy.

\clearpage
\bibliography{aaai25}
\clearpage
\appendix
\section{Appendix} 
\subsection{Impact Statements} 
\label{appendix_impact}
Our work provides a both cost-efficient and high-performance framework for solving reasoning problems with LLMs, which is expected to benefit broad organizations, such as the NLP research community and industrial companies. By designing such a cost-efficient framework, we empower these organizations to harness the reasoning ability of LLMs conveniently, especially for reasoning problems with high complexity.  
Our work not only makes financial savings but also benefits sustainability development by reducing the carbon emissions brought by extensive computation of running LLMs.



\subsection{Comparison with existing works}
In this section, we review some existing frameworks of LLM reasoning to clarify their difference with our method.
We use $p$, $s$, and $f(\cdot)$ to denote the reasoning problem, solution and the used LLM respectively. 

\noindent
\textbf{Chain-of-Thought (CoT)} 
CoT enhances LLM reasoning abilities by instructing the model to conduct step-by-step thinking: $p \rightarrow z_1,  \cdot \cdot \cdot, z_t \rightarrow s$, where $z_1, \cdot \cdot \cdot, z_t$ are intermediate thoughts during reasoning. In each reasoning step, thoughts are generated from a single LLM, which is formulated as follows: 
\begin{equation}
   z_n= f(p;\{z_m|m<n\}).
\end{equation}
\textbf{Tree-of-Thoughts (ToT)}
ToT uses the LLM to deliberate on
multiple reasoning paths and make high-quality global decisions via tree search. Formally, at the n-th reasoning step, ToT generates $N$ thought candidates from a single LLM: 
\begin{equation}
    \{z_n^k|k=1,2, \cdot \cdot \cdot, N\} = f(p;\{z_m|m<n\}).
\end{equation} 
Then it evaluates all candidate thoughts and selects the best one as the final thought $z_n$ at step $n$.

The above two well-known methods can enhance LLM reasoning abilities but are limited to using a single LLM (either small-scale or large-scale), suffering from either low performance or high token cost issues.
As for most multi-agent debate methods \cite{liang2023encouraging,du2023improving,wang2023unleashing}, they are also focusing on reasoning with larger-scale LLMs, resulting in challenges on complex reasoning tasks due to the expensive API costs. 
To address this issue, we propose an adaptive synergy framework composed of hybrid LLMs, fully exploiting the unique strengths of different-scale LLMs.

\subsection{Token Prices of Used LLMs} \label{appendix_cost}
Here we show the token prices of different LLMs we use in this work in Table \ref{cost}. 
The statistics of GPT-3.5 and GPT-4 are from the official report of OpenAI\footnote{https://openai.com/pricing}. 
The statistics of Yi-34B are from the official report of 01.AI\footnote{https://platform.lingyiwanwu.com/}.
The statistics of Gemini1pro are from the official report of Google\footnote{https://ai.google.dev/pricing}.
The statistics of Mistral-7B are from the official report of Mistral AI\footnote{https://mistral.ai/technology}.
The statistics of LLaMA-13B are from Baidu online service platform\footnote{https://console.bce.baidu.com/qianfan/ais/console/onlineService}.
PaLM2 is cost-free when the work is done.
\begin{table}[ht!]
    \centering
    \begin{tabular}{lll}
    \hline
        Model & Input /1M tokens & Output /1M tokens \\ \hline
        GPT-3.5 & \$1.5 & \$2 \\ 
        GPT-4 & \$30 & \$60 \\ 
        Yi-34B & \$0.35 & \$0.35 \\ 
        Gemini1pro & \$0.5 & \$1.5 \\
        Mistrial-7B & \$0.25  &\$0.25 \\ 
        LLaMA-13B & \$0.28 & \$0.28 \\ 
        PaLM2 & 0 & 0 \\  \hline
    \end{tabular}
    \caption{Input and output token prices of used LLMs.}
    \label{cost}
\end{table}

\subsection{Specific Algorithm of System 1} 
\label{appendix_alg1}
We show the specific algorithm for implementing System 1 in Algorithm 1.
\begin{algorithm}[ht!] 
\caption{Algorithm of System 1} 
\label{alg1}
\KwIn {Reasoning task description $p$, thoughts of the last step $a$, the number of hybrid LLMs $K$} 
\For{$k \in \{1,2,...,K\}$}
{$ a_k^{(1)} = f_{Ik}(p;a)$ \tcp{Generate initial thoughts}}
\For{$j \in \{1,2,...,K\}$}{
\For{$k \in \{1,2,...,K\} \setminus \{j\}$}
{$ h_{j \rightarrow k}^{(2)} = f_{Ij}(p_{inter};a_k^{(1)})$ \tcp{Multiple-intuition interactions}}}
\For{$k \in \{1,2,...,K\}$}{
$ a_{k}^{(3)} = f_{Ik}(p_{update};\sum\limits_{j \in \{1,2,...,K\} \setminus \{k\}} h_{j \rightarrow k}^{(2)})$ \tcp{Update intuitions}}

$H = \{a_1^{(3)}, a_2^{(3)}, ..., a_K^{(3)}\}$ \\
\Return $H$
\end{algorithm}

\subsection{Specific Algorithm of SoT} \label{appendix_alg2}
We show the specific algorithm of the whole framework of SoT in Algorithm 2.
\begin{algorithm}[ht!] 
\caption{Algorithm of SoT} 
\label{alg2}
\KwIn{Required reasoning steps $N$, task description prompt in each reasoning step $\{p_0, ..., p_N\}$} 
$t = 0$ \tcp{Current reasoning step}
$a_t = None$ \tcp{Intialize current thoughts}
\While{$t \leq N$}{
$t = t+1$ \\
$H_t =  System \; 1(p_t;a_{t-1})$ \tcp{Propose intuitions by System 1}
$p, a_t = Confidence \; Evaluator (H_t)$ \tcp{Confidence evaluation}
\If p 
{
$a_t = System \; 2(p_{ref};a_t)$ \tcp{{Intervention with reflective System 2}}
}
}
\Return $a_t$
\end{algorithm}

\subsection{Empirical Average Intervention Rates} \label{appendix_emp}
Here we show the empirical average intervention rates of different LLM combinations in System 1 on six tasks in Table \ref{reject}, for measuring the practicability to maintain cost-saving by using SoT.
\begin{table*}[!t]
\setlength{\tabcolsep}{1mm}
    \begin{tabular}{lllllll}
    \hline
        LLM combinations & Game of 24 & Logic Grid Puzzle & GSM8K &  Creative Writing & OpenQA & Constrained Generation \\ \hline
        3 GPT-3.5  &28\% & 49\% & 24\% & 51\% & 55\% & 56\% \\
        3 PaLM2  &33\% & 55\% & 27\% & 57\% & 57\% & 61\% \\
        3 Gemini1pro  &30\% & 53\% & 26\% & 50\% & 54\% & 58\% \\
        GPT-3.5/PaLM2/Gemini1pro  &26\%  &44\% &23\%  &42\% &52\% &53\% \\ 
        3 LLaMA-13B   &39\% & 68\% & 41\% & 65\% & 67\% & 65\% \\
        3 Mistral-7B  &36\% & 61\% & 39\% & 60\% & 61\% & 63\% \\
        3 Yi-34B   &36\% & 59\% & 36\% & 61\% & 59\% & 59\% \\ 
        LLaMA-13B/Mistral-7B/Yi-34B &33\% & 54\% & 35\% & 49\% & 54\% & 57\% \\
        \hline
    \end{tabular}
    \caption{Empirical average intervention rate on six reasoning tasks.}
    \label{reject}
\end{table*}

\subsection{Supplement of Main Results} \label{appendix_main}
Here we supplement the main results about $SoT_C$ on Constrained Generation and Open-ended QA tasks in Table \ref{tab:main7} and Table \ref{tab:main8}. The results show that $SoT_C$ outperforms all baselines, consistent with the conclusion in the main text.
\begin{table}[t!]
\setlength{\tabcolsep}{1mm}
    \begin{tabular}{lllll}
    \hline
        Methods & FairEval & Diversity & Cost (\$) & TFLOPS \\ \hline
        $SoT_C$ & ~ & \textbf{6.2} & 8.32 & 5.44E+05 \\ \hline
        v.s. CoT(best of 1) & 73.0\% & 4.2 & 4.84 & 2.91E+05 \\
        v.s. CoT(best of 5)  & 66.4\% & 4.2 & 20.11 & 1.21E+06 \\
        v.s. Self-refine & 59.2\% & 5.3 & 31.86 & 1.91E+06 \\ 
        v.s. ToT & 62.9\% & 4.7 & 58.52 & 3.51E+06 \\ 
        v.s. SPP & 70.7\% & 4.6 & 26.69 & 1.60E+06 \\ 
        v.s. MAD+judge & 57.6\% & \underline{5.6} & 36.59 & 2.20E+06 \\ \hline
    \end{tabular}
    \caption{Results on Constrained Generation task of $SoT_C$ (FairEval value larger than 50\% means results from $SoT_C$ are better).}
    \label{tab:main7}
\end{table}

\begin{table}[t!]
\setlength{\tabcolsep}{1mm}
\begin{tabular}{lllll}
    \hline
        Methods & FairEval & Diversity & Cost (\$) & TFLOPS \\ \hline
        $SoT_C$ & ~ & \textbf{5.2} & 6.77 & 4.72E+05 \\ \hline
        v.s. CoT(best of 1) & 68.4\% & 3.1 & 2.27 & 1.36E+05 \\ 
        v.s. CoT(best of 5) & 62.8\% & 3.3 & 8.72 & 5.23E+05 \\
        v.s. Self-refine & 60.5\% & 4.2 & 15.27 & 9.16E+05 \\ 
        v.s. ToT & 62.3\% & 3.3 & 19.44 & 1.17E+06 \\ 
        v.s. SPP & 64.5\% & 3.8 & 8.33 & 5.00E+05 \\ 
        v.s. MAD+judge & 58.1\% & \underline{4.7} & 17.00 & 1.02E+06 \\  \hline
    \end{tabular}
    \caption{Results on Open-ended Question Answer task (FairEval value larger than 50\% means results from $SoT_C$ are better).}
    \label{tab:main8}
\end{table}

\subsection{Statistical Tests of Results}
Here we report the reasoning accuracy of SoT with standard error in five independent trials in Table~\ref{tab:se}.
It can be seen that SoT achieves significant performance improvement compared with baselines.

\begin{table*}[t!]
    \centering
    \begin{tabular}{ccccc}
    \hline
    Method & Game of 24 & Trivia Creative Writing & Logic Grid Puzzle & GSM8K \\ \hline
        CoT(best of 1) & 4\% & 67.1\%  & 65.8\% & 87.8\%  \\ 
        CoT(best of 5) & 14\% & 73.4\% & 67.1\% & 91.3\%  \\ 
        LLM-cascade & 8\% & 65.7\% & 62.1\% & 89.1\%  \\ 
        Self-refine & 20\% & 78.2\% & 60.6\% & 91.1\% \\ 
        ToT & 64\% & 76.8\% & 66.1\% & 91.8\%  \\ 
        SPP & 12\% & 79.9\% & 68.3\% & 84.6\% \\ 
        MAD+judge & 22\% & 77.4\% & 66.8\% & 89.3\% \\ \hline
        $SoT_O$ & \underline{73±2\%} & \underline{82.2±0.7\%} & \underline{69.9±0.8\%} & \underline{93.4±0.3\%} \\ 
            $SoT_C$ & \textbf{76±1\%} & \textbf{83.1±0.9\%}  & \textbf{71.5±0.5\%}  & \textbf{94.0±0.5\%}  \\ \hline
    \end{tabular}
    {\caption{Performance comparison with standard error on Game of 24, Trivia Creative Writing, Logic Grid Puzzle and GSM8K tasks.}
    \label{tab:se}}
\end{table*}

\subsection{Performance-cost Trade-off Analysis on More Reasoning Tasks} \label{appendix_tradeoff}
\begin{figure*}[!t]
\begin{center}
\centerline{\includegraphics[width=\textwidth]{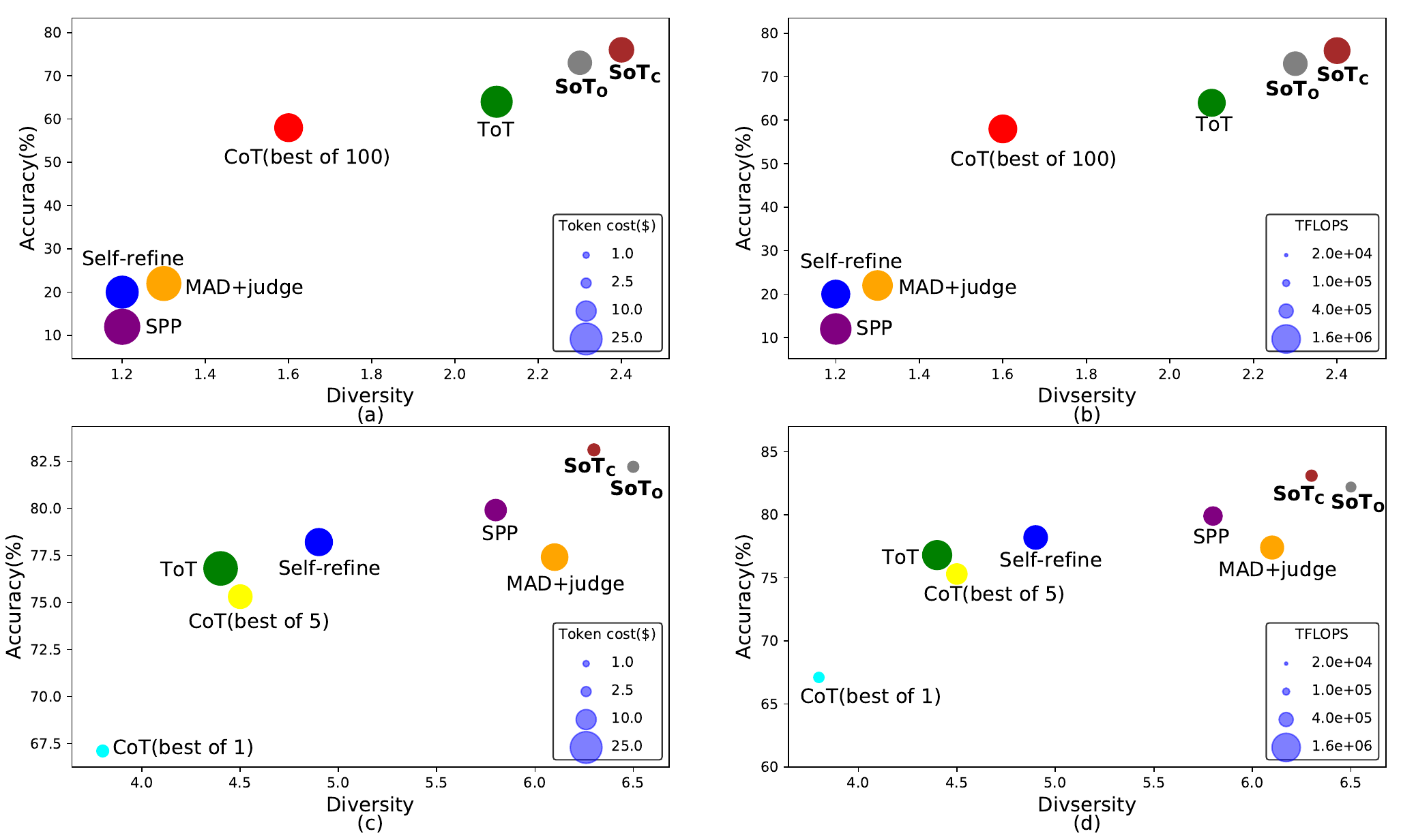}}
\caption{The reasoning accuracy, solution diversity versus token costs/TFLOPS on Game of 24 (a) (b) and Trivia Creative Writing (c) (d). SoT achieves a better performance-cost trade-off than all compared methods.}
\label{scatter}
\end{center}
\end{figure*}

\begin{figure*}[!ht]
\begin{center}
\includegraphics[width=\textwidth]{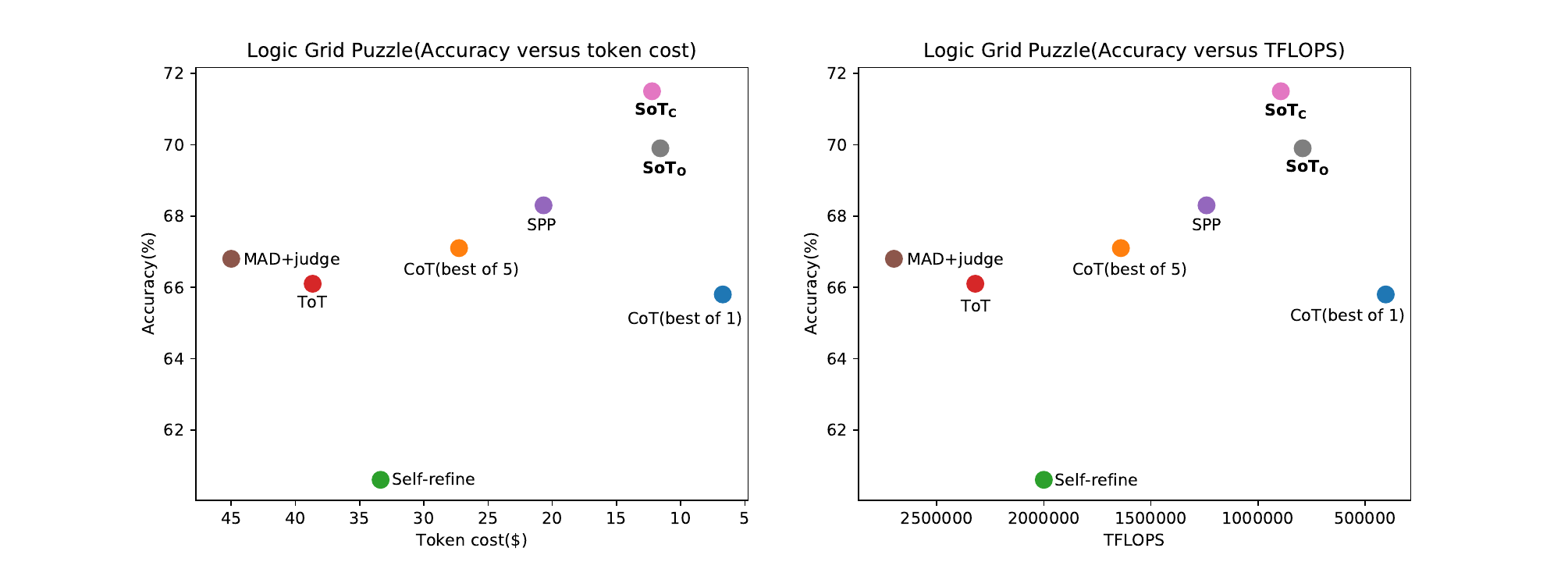}
\caption{The reasoning accuracy versus token costs/TFLOPS on Logic Grid Puzzle task. SoT achieves a better performance-cost trade-off than all compared methods.}
\label{scatter1}
\end{center}
\end{figure*}

\begin{figure*}[!ht]
\begin{center}
\includegraphics[width=\textwidth]{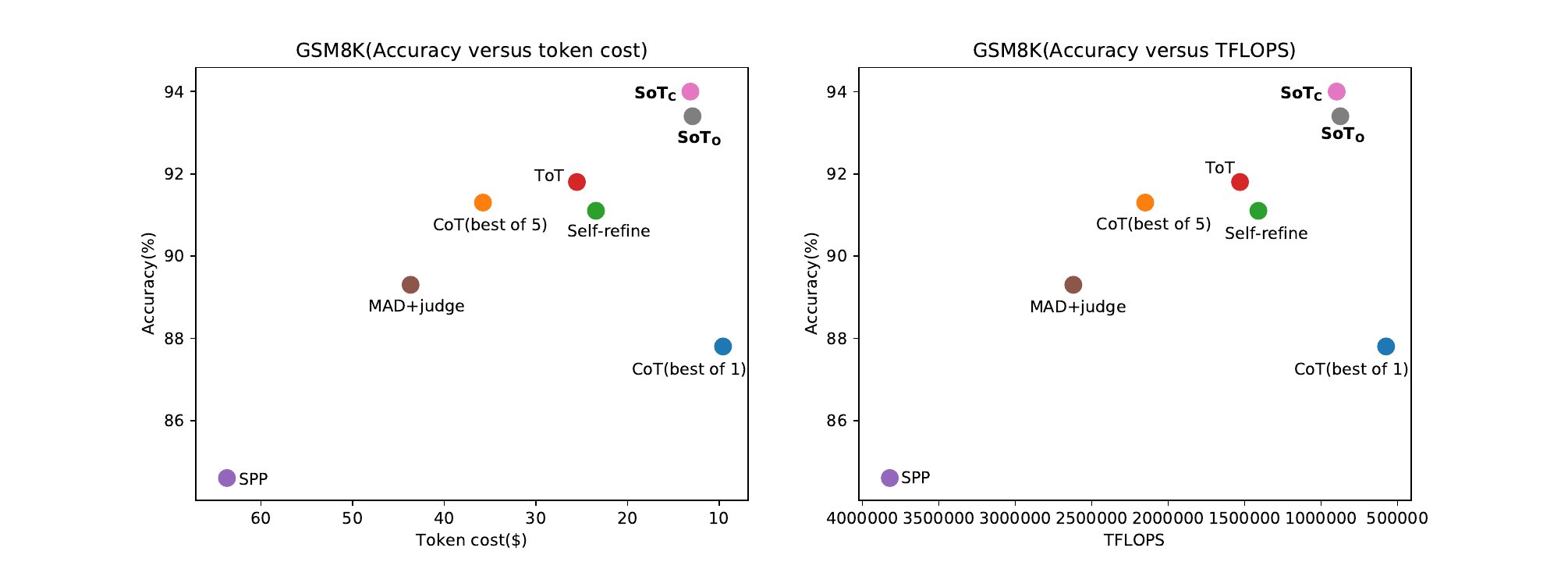}
\caption{The reasoning accuracy versus token costs/TFLOPS on GSM8K task. SoT achieves a better performance-cost trade-off than all compared methods.}
\label{scatter2}
\end{center}
\end{figure*}

\begin{figure*}[!ht]
\begin{center}
\includegraphics[width=\textwidth]{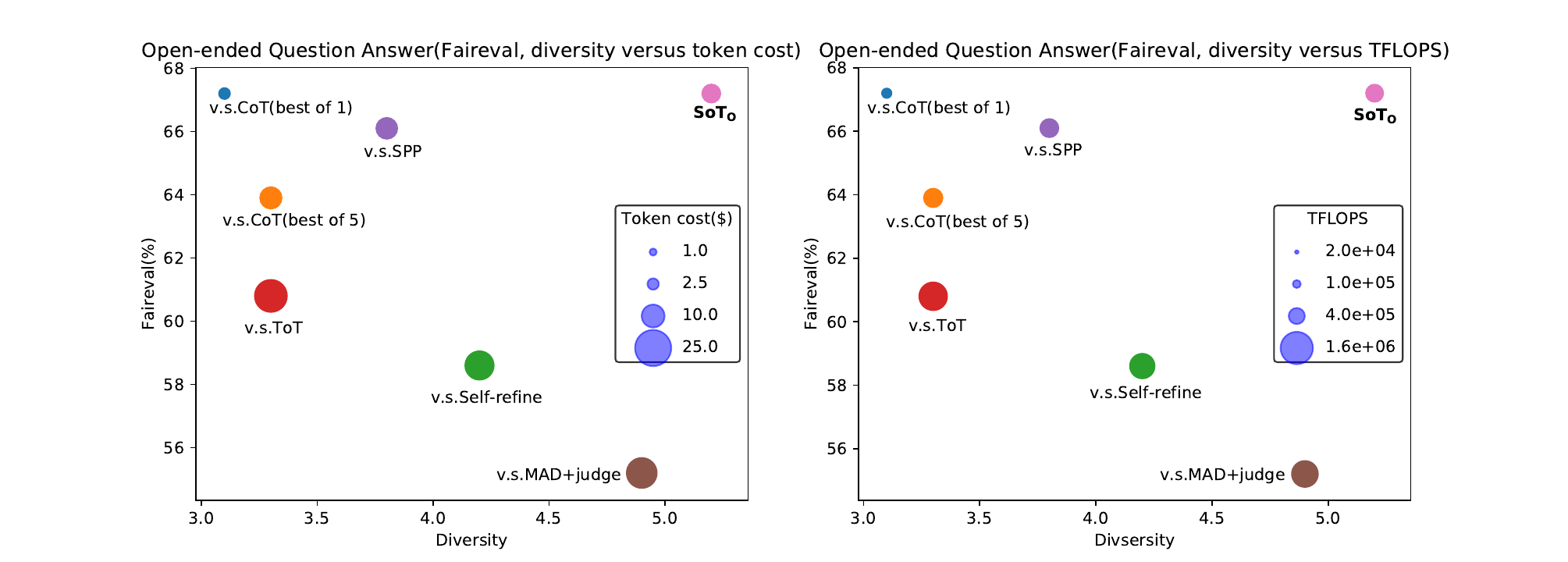}
\caption{The reasoning accuracy, solution diversity versus token costs/TFLOPS of $SoT_O$ and baselines on Open-ended QA task. $SoT_O$ achieves a better performance-cost trade-off than all compared methods.}
\label{scatter3}
\end{center}
\end{figure*}

\begin{figure*}[!ht]
\begin{center}
\includegraphics[width=\textwidth]{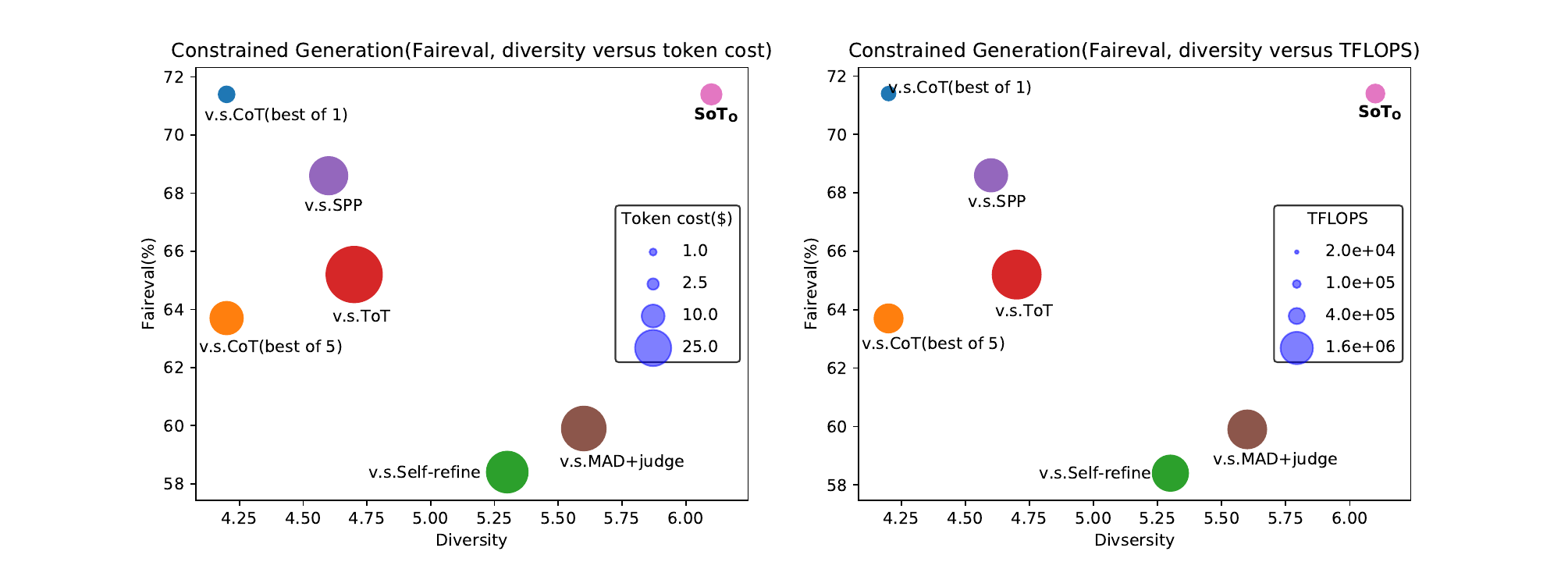}
\caption{The reasoning accuracy, solution diversity versus token costs/TFLOPS of $SoT_O$ and baselines on Constrained Generation task. $SoT_O$ achieves a better performance-cost trade-off than all compared methods.}
\label{scatter4}
\end{center}
\end{figure*}

\begin{figure*}[!ht]
\begin{center}
\includegraphics[width=\textwidth]{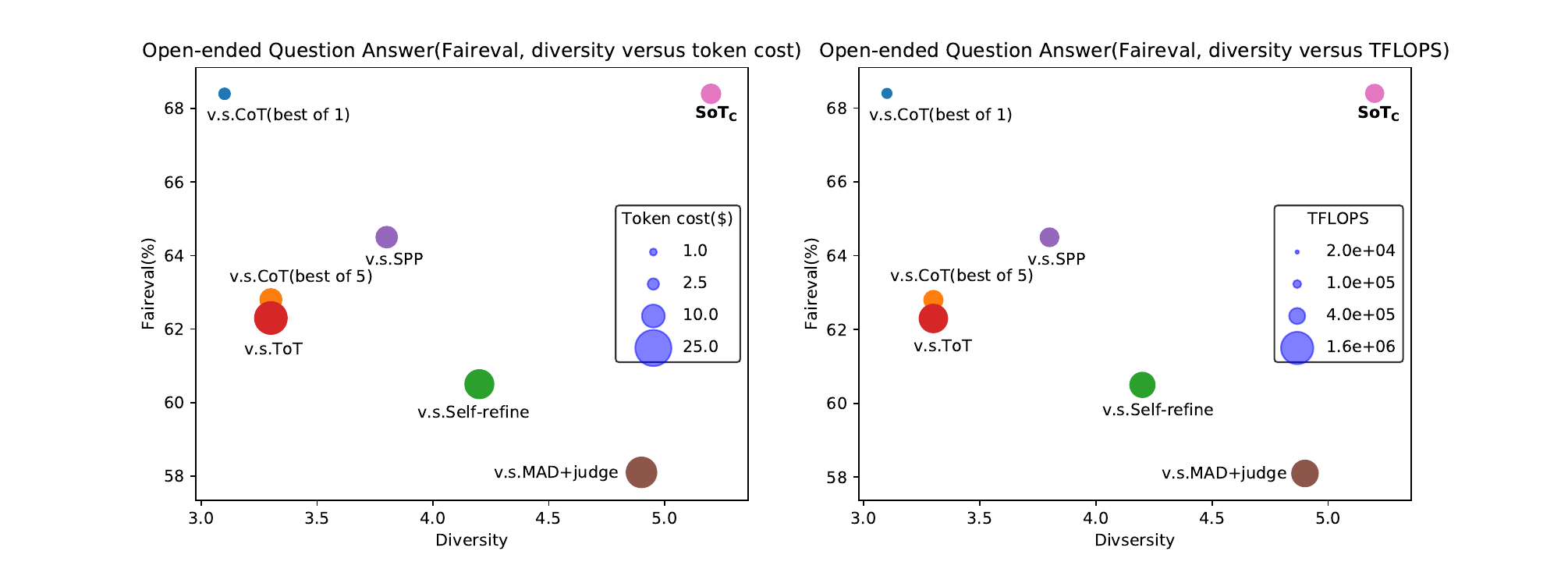}
\caption{The reasoning accuracy, solution diversity versus token costs/TFLOPS of $SoT_C$ and baselines on Open-ended QA task. $SoT_C$ achieves a better performance-cost trade-off than all compared methods.}
\label{scatter5}
\end{center}
\end{figure*}

\begin{figure*}[!ht]
\begin{center}
\includegraphics[width=\textwidth]{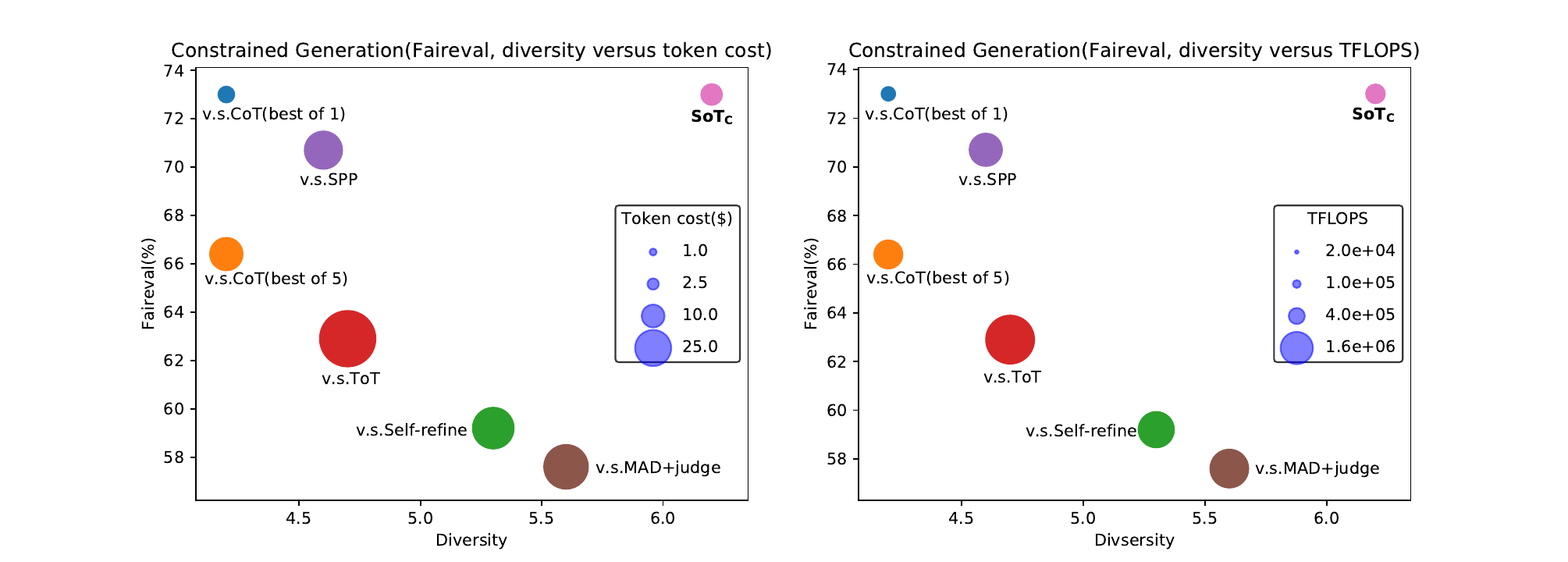}
\caption{The reasoning accuracy, solution diversity versus token costs/TFLOPS of $SoT_C$ and baselines on Constrained Generation task. $SoT_C$ achieves a better performance-cost trade-off than all compared methods.}
\label{scatter6}
\end{center}
\end{figure*}

We conduct performance-cost visualization analysis on all reasoning tasks. The results are shown in Figure \ref{scatter}, \ref{scatter1}, \ref{scatter2}, \ref{scatter3}, \ref{scatter4}, \ref{scatter5} and \ref{scatter6}. Consistent with the results in the main text, SoT achieves the best performance-cost trade-off among all methods.

\subsection{Illustrations of SoT on More Reasoning Tasks} \label{appendix_case}
\begin{figure*}[!ht]
\begin{center}
\centerline{\includegraphics[width=1\textwidth]{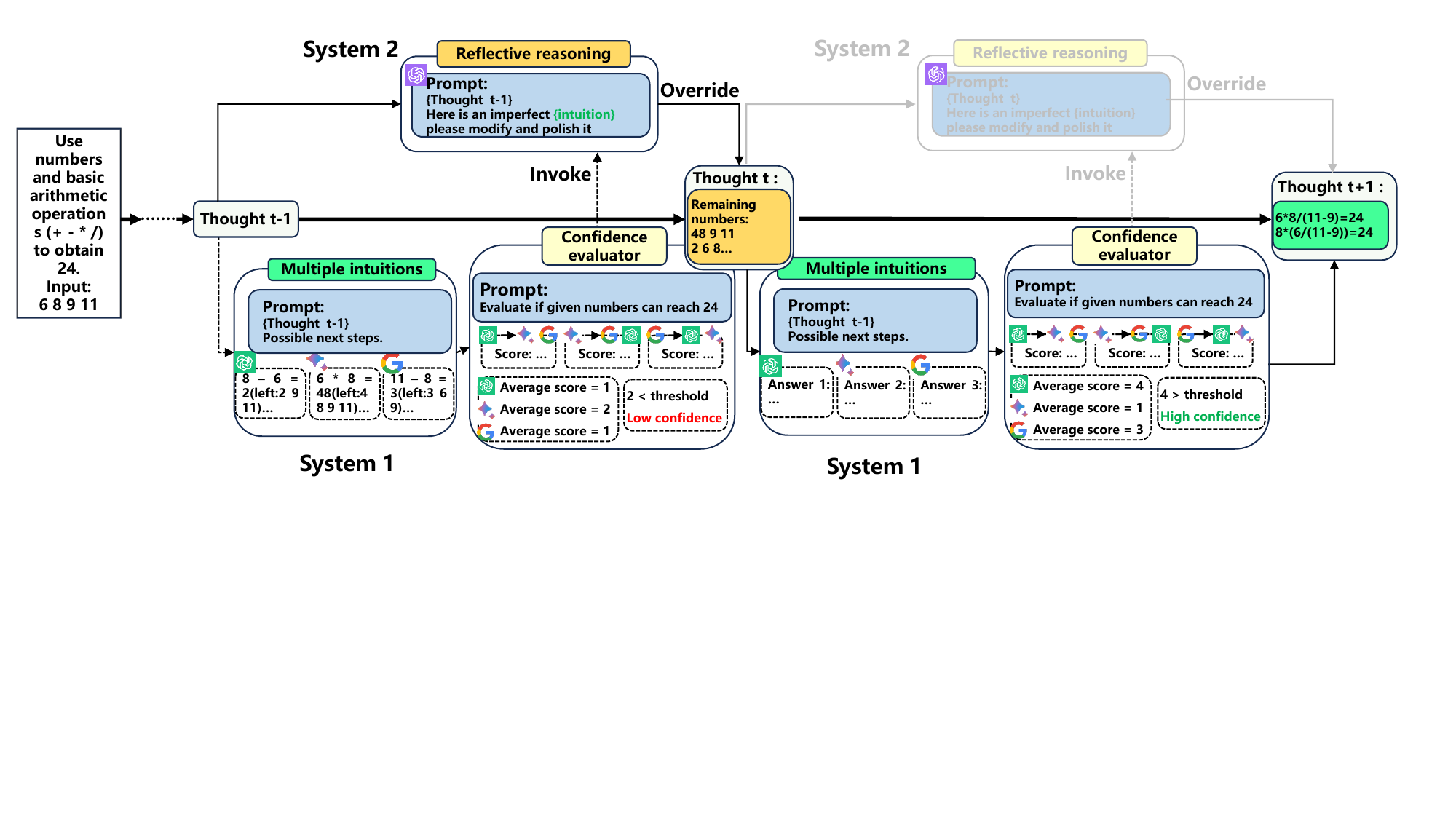}}
\caption{An illustrative example of SoT from Game of 24 Task.}
\label{case1}
\end{center}
\vskip -0.8in
\end{figure*}

\begin{figure*}[!ht]
\begin{center}
\centerline{\includegraphics[width=1\textwidth]{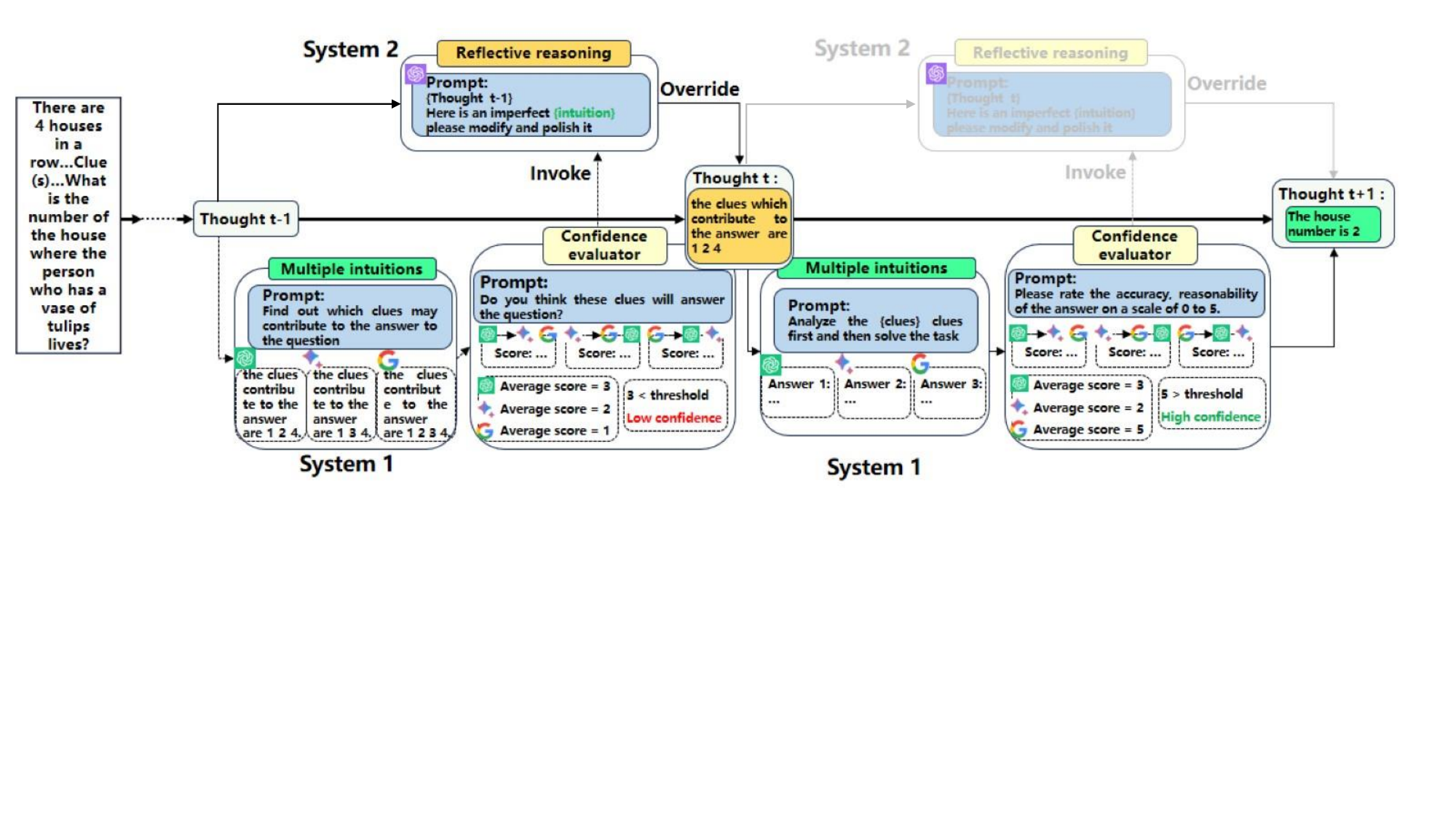}}
\caption{An illustrative example of SoT from Logic Grid Puzzle Task.}
\label{case2}
\end{center}
\vskip -0.8in
\end{figure*}

\begin{figure*}[!ht]
\begin{center}
\centerline{\includegraphics[width=1\textwidth]{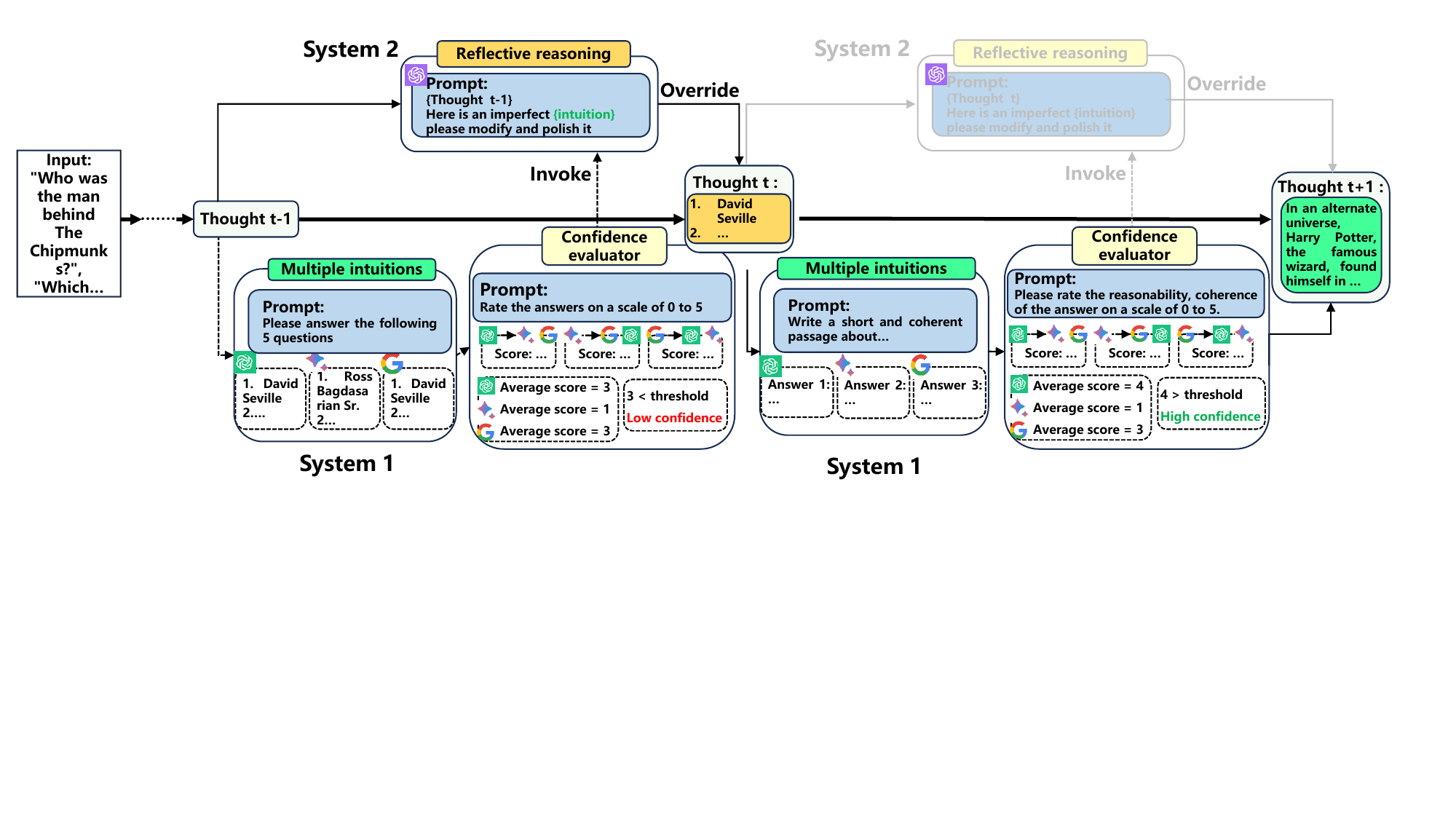}}
\caption{An illustrative example of SoT from Trivia Creative Writing Task.}
\label{case3}
\end{center}
\vskip -0.8in
\end{figure*}

\begin{figure*}[!ht]
\begin{center}
\centerline{\includegraphics[width=1\textwidth]{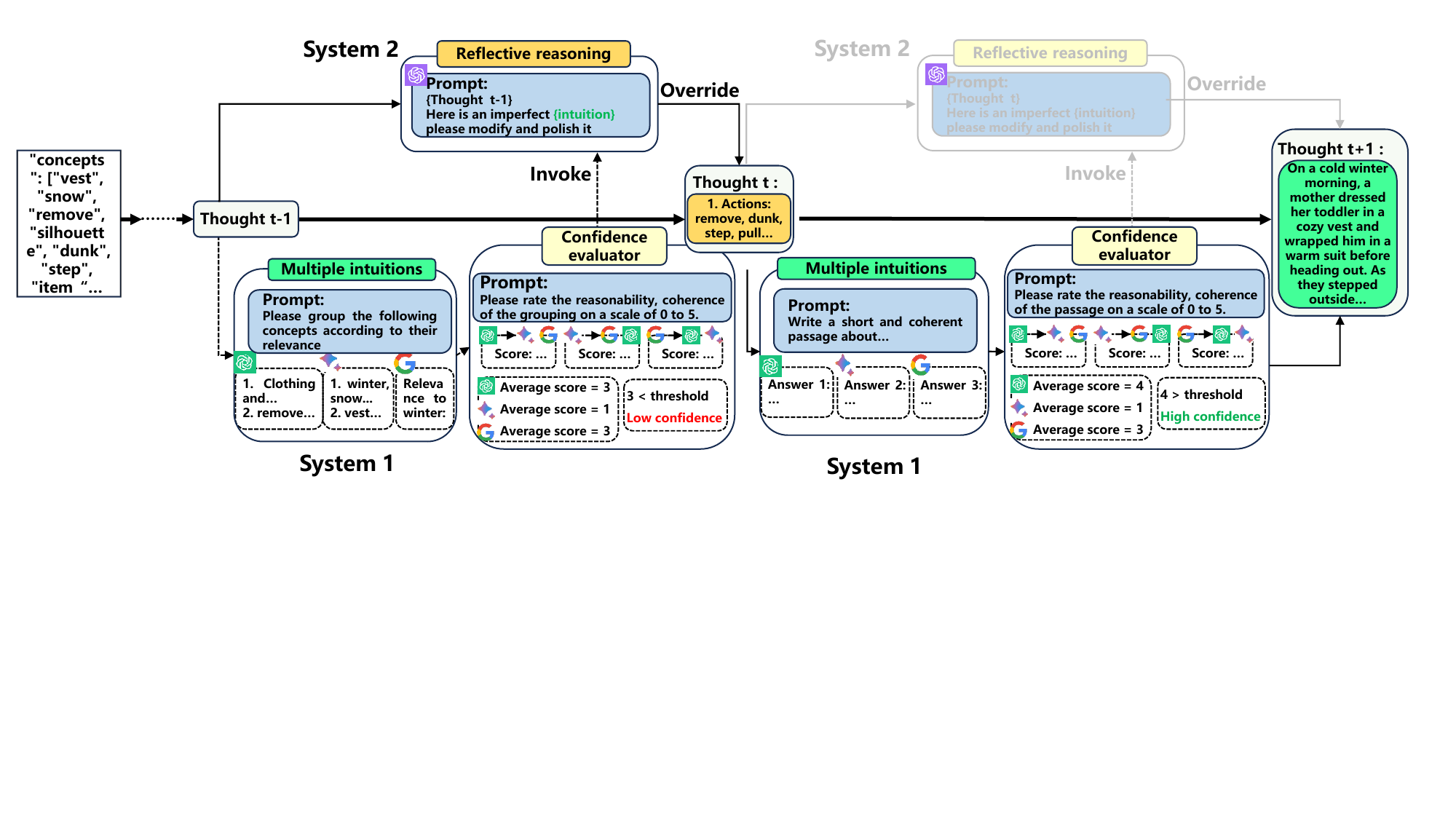}}
\caption{An illustrative example of SoT from Constrained Generation Task.}
\label{case4}
\end{center}
\vskip -0.8in
\end{figure*}

In the main paper, we illustrate SoT with an example from Open-ended QA tasks. For a better understanding of the scheme of SoT, here we present more cases from some other reasoning tasks in Figure \ref{case1}, \ref{case2}, \ref{case3} and \ref{case4}.

\subsection{Prompts used in SoT}
\begin{figure*}[t!]
\begin{center}
\centerline{\includegraphics[width=0.9\textwidth]{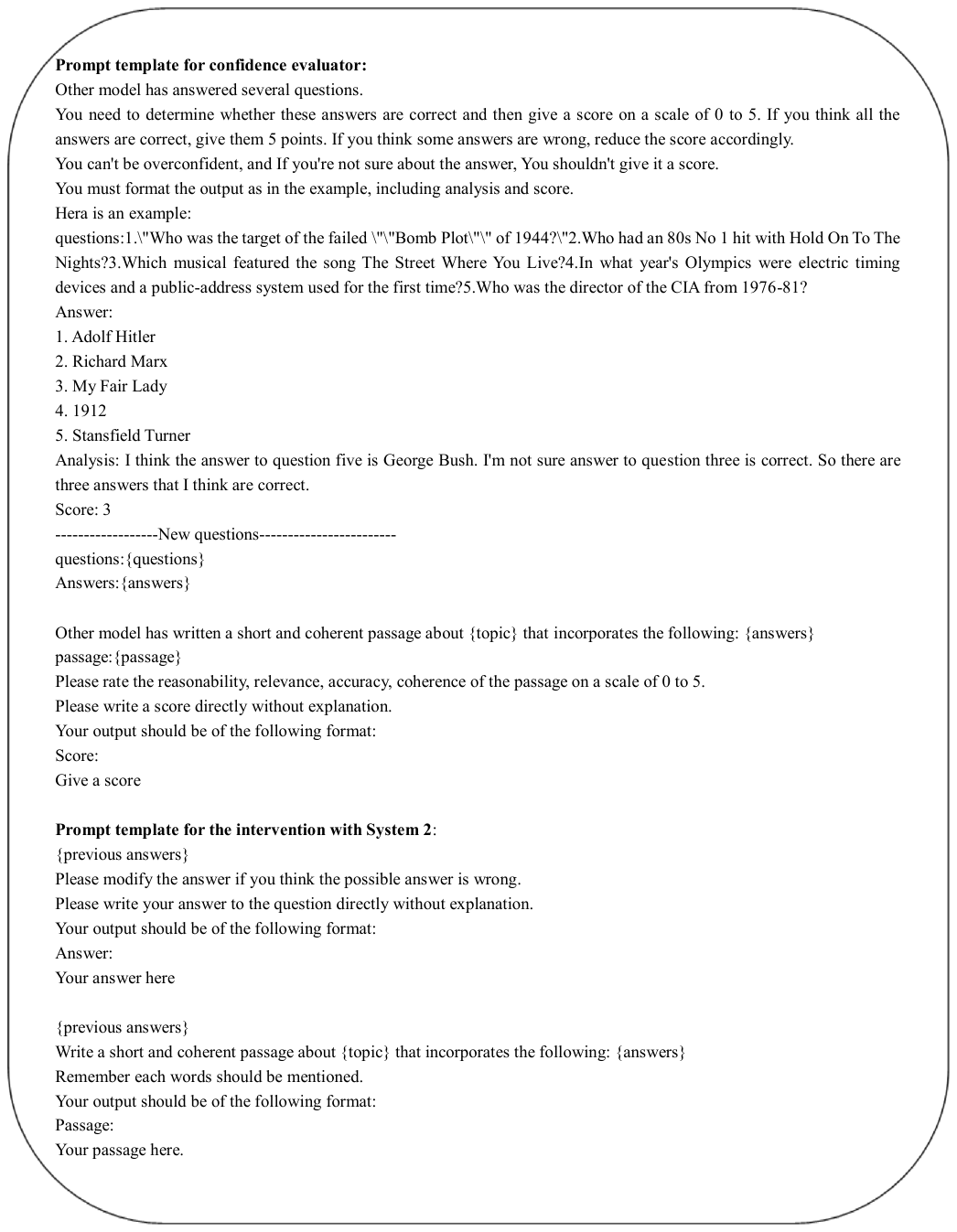}}
\caption{Prompts of the confidence evaluator and intervention with System 2 in SoT on Trivia Creative Writing Task.}
\label{prompt1}
\end{center}
\end{figure*}

\begin{figure*}[t!]
\begin{center}
\centerline{\includegraphics[width=0.9\textwidth]{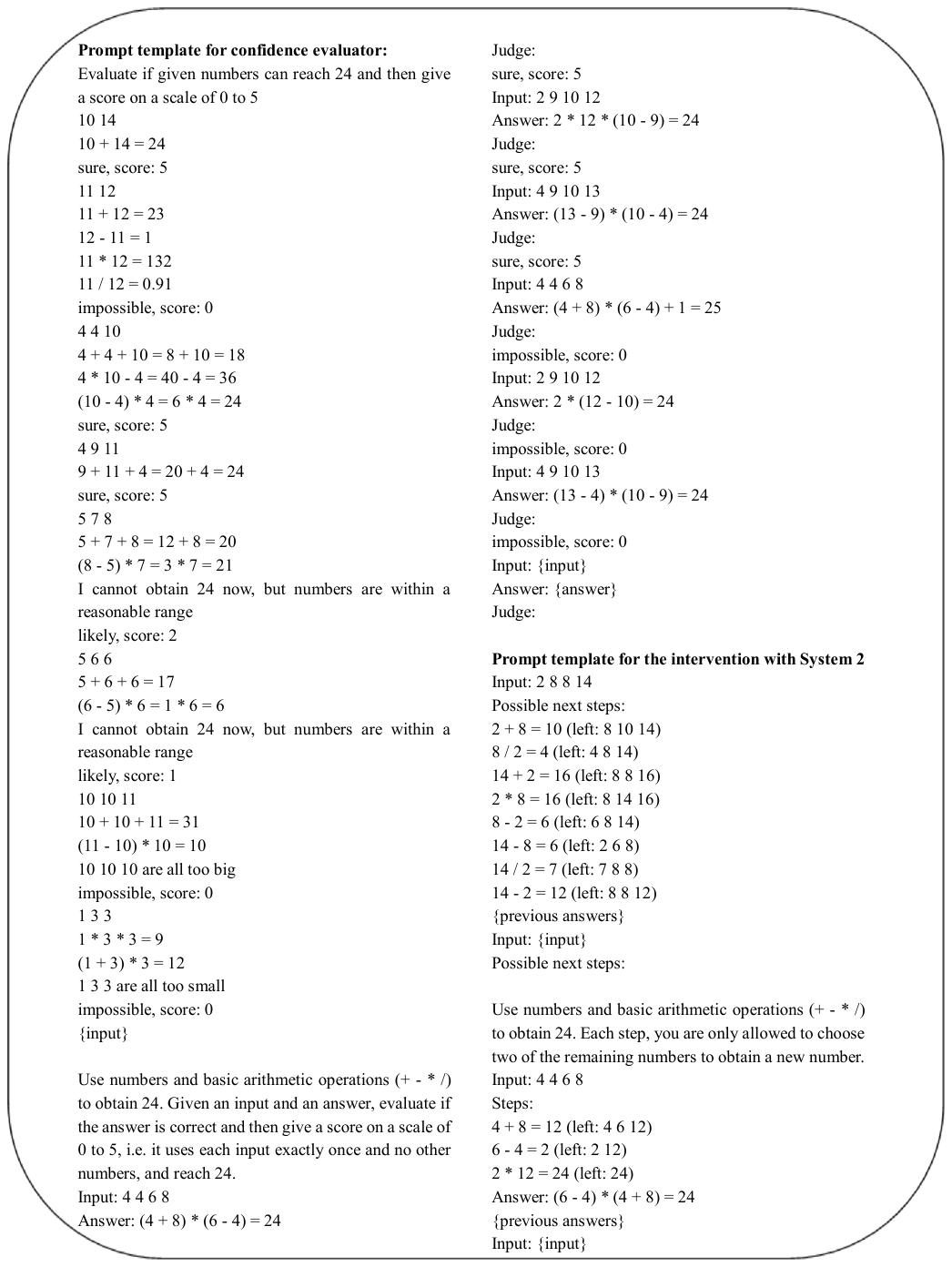}}
\caption{Prompts of the confidence evaluator and intervention with System 2 in SoT on Game of 24 Task.}
\label{prompt2}
\end{center}
\end{figure*}

We present the designed prompt templates for the confidence evaluator and intervention with System 2 when implementing SoT, taking Trivia Creative Writing (see Figure~\ref{prompt1}) and Game of 24 (see Figure~\ref{prompt2}) tasks as examples. 



\end{document}